\documentclass[11pt,a4paper]{article}
\pdfoutput=1

\usepackage[numbers,sort&compress]{natbib}
\usepackage{sjtu_iic_arxiv}
\usepackage{mathtools}
\usepackage{nicefrac}
\usepackage{etoolbox}
\usepackage{float}
\usepackage{multirow}
\usepackage{longtable}

\newtheorem{definition}{Definition}

\newcommand{\xinyuntodo}[1]{}

\newcommand{\bestval}[1]{\textbf{#1}}
\newcommand{\secondval}[1]{\underline{#1}}

\DeclareRobustCommand{\textttt}[1]{{\fontencoding{T1}\bfseries\fontshape{sc}\selectfont #1}}
\newcommand{\tasktext}[1]{\textsc{#1}\xspace}
\newcommand{\taskheading}[1]{\textttt{#1.}}
\newcommand{\taskwithsymbol}[2]{\textttt{#1} (#2).}
\newcommand{\taskbold}[1]{\textttt{#1}}
\newcommand{\build}{\tasktext{Build}}
\newcommand{\revise}{\tasktext{Revise}}
\newcommand{\explain}{\tasktext{Explain}}

\newcommand{\buildtask}{\ensuremath{\mathcal{T}_{\text{\textsc{Build}}}}}
\newcommand{\revisetask}{\ensuremath{\mathcal{T}_{\text{\textsc{Revise}}}}}
\newcommand{\explaintask}{\ensuremath{\mathcal{T}_{\text{\textsc{Explain}}}}}

\newcommand{\workspacesym}{\ensuremath{\mathcal{W}}}
\newcommand{\docset}{\ensuremath{\mathcal{D}}}
\newcommand{\paramset}{\ensuremath{\mathcal{P}}}
\newcommand{\codeset}{\ensuremath{\mathcal{S}}}
\newcommand{\envset}{\ensuremath{\mathcal{E}}}
\newcommand{\metricset}{\ensuremath{\mathcal{M}}}
\newcommand{\workspacetuple}{\ensuremath{\langle \docset,\; \paramset,\; \codeset,\; \envset,\; \metricset \rangle}}
\newcommand{\workspacedef}{\ensuremath{\workspacesym = \workspacetuple}}

\newcommand{\objmetric}{\ensuremath{\mathcal{M}_{\mathrm{obj}}}}
\newcommand{\expmetric}{\ensuremath{\mathcal{M}_{\mathrm{exp}}}}
\newcommand{\agentobj}{\ensuremath{v}}
\newcommand{\refobj}{\ensuremath{v^\star}}
\newcommand{\objtol}{\ensuremath{\epsilon}}
\newcommand{\indicator}{\ensuremath{\mathbf{1}}}
\newcommand{\passdelta}{\ensuremath{\Delta_{\mathrm{R-B}}}}
\newcommand{\fsflatdelta}{\ensuremath{\Delta_{\mathrm{Fs-Flat}}}}
\newcommand{\rankcorr}{\ensuremath{\rho}}

\newcommand{\pulpmodel}{\texttt{pulp.LpProblem}\xspace}
\newcommand{\revisecode}{\textsc{Revise-code}\xspace}
\newcommand{\revisemodel}{\textsc{Revise-model}\xspace}
\newcommand{\reviseall}{\textsc{Revise-all}\xspace}
\newcommand{\fsvariant}{\textsc{Fs}\xspace}
\newcommand{\flatvariant}{\textsc{Flat}\xspace}
\newcommand{\objavg}{\textsc{Obj. Avg}\xspace}

\newcommand{\optimalstatus}{\textsc{Optimal}\xspace}
\newcommand{\wrongvalue}{\textsc{WrongValue}\xspace}
\newcommand{\runtimeerror}{\textsc{RuntimeError}\xspace}
\newcommand{\emptyoutput}{\textsc{EmptyOutput}\xspace}
\newcommand{\apiexception}{\textsc{ApiException}\xspace}
\newcommand{\typeerror}{\textsc{TypeError}\xspace}
\newcommand{\nameerror}{\textsc{NameError}\xspace}
\newcommand{\taubench}{\ensuremath{\tau}\nobreakdash-bench\xspace}

\definecolor{SJTUIICGroup}{HTML}{EAF5F8}
\makeatletter
\newcommand{\SJTUIICStartTableLines}{\@ifundefined{nolinenumbers}{}{\nolinenumbers}}
\newcommand{\SJTUIICEndTableLines}{\@ifundefined{linenumbers}{}{\linenumbers}}
\newcommand{\sjtuiicgroup}[2]{\@nameuse{sjtuiicgroup@#1}{#2}}
\@namedef{sjtuiicgroup@3}#1{\rowcolor{SJTUIICGroup}\emph{#1} & &\\}
\@namedef{sjtuiicgroup@4}#1{\rowcolor{SJTUIICGroup}\emph{#1} & & &\\}
\@namedef{sjtuiicgroup@5}#1{\rowcolor{SJTUIICGroup}\emph{#1} & & & &\\}
\@namedef{sjtuiicgroup@6}#1{\rowcolor{SJTUIICGroup}\emph{#1} & & & & &\\}
\@namedef{sjtuiicgroup@8}#1{\rowcolor{SJTUIICGroup}\emph{#1} & & & & & & &\\}
\makeatother
\AtBeginEnvironment{table}{\SJTUIICStartTableLines}
\AtEndEnvironment{table}{\SJTUIICEndTableLines}
\newif\ifarxivversion
\arxivversiontrue
\newenvironment{promptbox}[1]{%
  \begin{tcolorbox}[title={#1}, colback=gray!5, colframe=black!50, breakable]
  \small\ttfamily
}{%
  \end{tcolorbox}
}

\hypersetup{
  pdftitle={OR-Space: A Full-Lifecycle Workspace Benchmark for Industrial Optimization Agents},
  pdfauthor={Chenyu Zhou, Xinyun Lu, Jiangyue Zhao, Jianghao Lin, Dongdong Ge, Yinyu Ye}
}

\AtBeginDocument{\renewcommand{\today}{2026-5-27}}
\setheadertext{%
  \raisebox{-0.55cm}{\includegraphics[height=1.3cm]{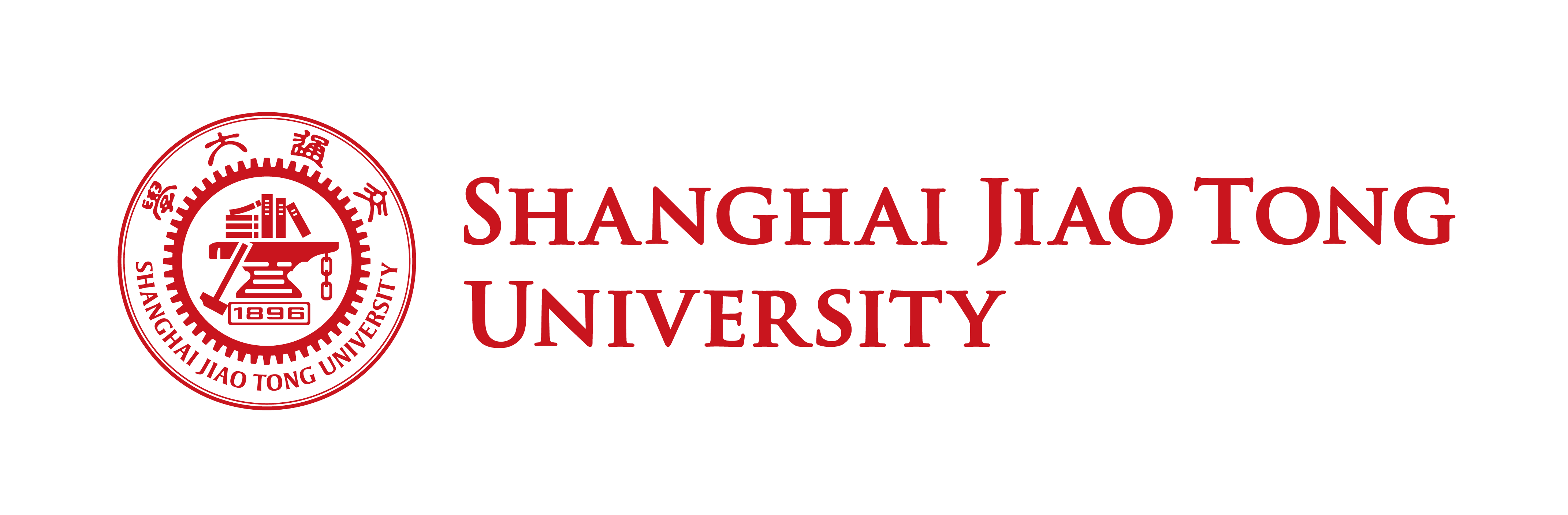}}%
  \hspace{0.15cm}%
  \raisebox{-0.29cm}{\includegraphics[height=0.72cm]{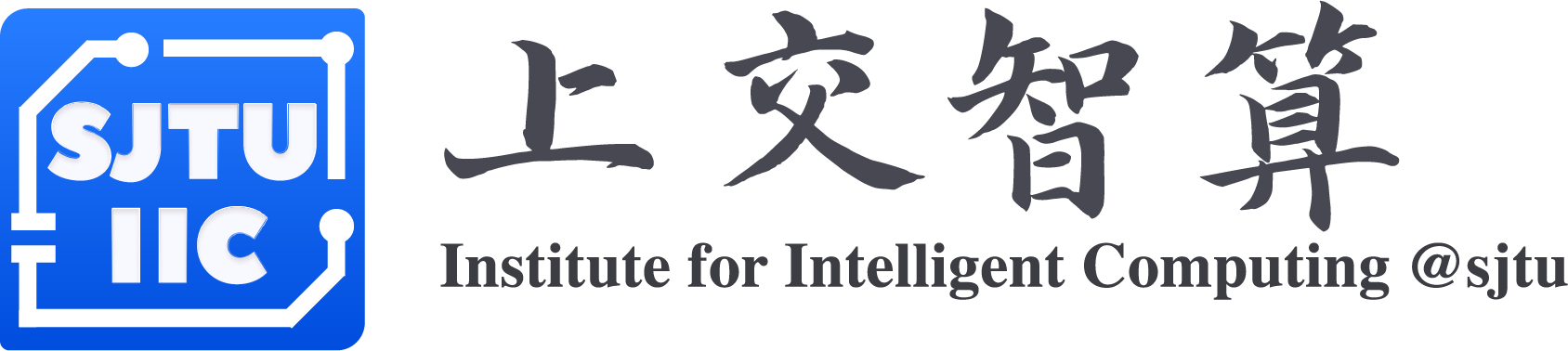}}%
}
\setheadertitle{OR-Space: A Full-Lifecycle Workspace Benchmark for Industrial Optimization Agents}
\githublink{https://github.com/0xzhouchenyu/OR-Space}
\huggingfacelink{https://huggingface.co/datasets/Chenyu-Zhou/OR-Space}
\correspondingemail{%
  {\fontsize{8.9}{10.7}\selectfont
  \makebox[\linewidth][l]{%
    \{chenyuzhou, linjianghao, ddge\}@sjtu.edu.cn\hspace{1.55em}%
    \{luxy, jiangyuezhao05\}@stu.sufe.edu.cn\hspace{1.55em}%
    yinyu-ye@stanford.edu}}\\[0.3cm]
  \textsuperscript{*}Equal contribution\quad
  \textsuperscript{\dag}Corresponding author
}

\newcommand{\sjtuiicTitleLogo}{%
   \raisebox{-0.25cm}{%
     \includegraphics[height=1.0cm,trim=250 170 250 170,clip]{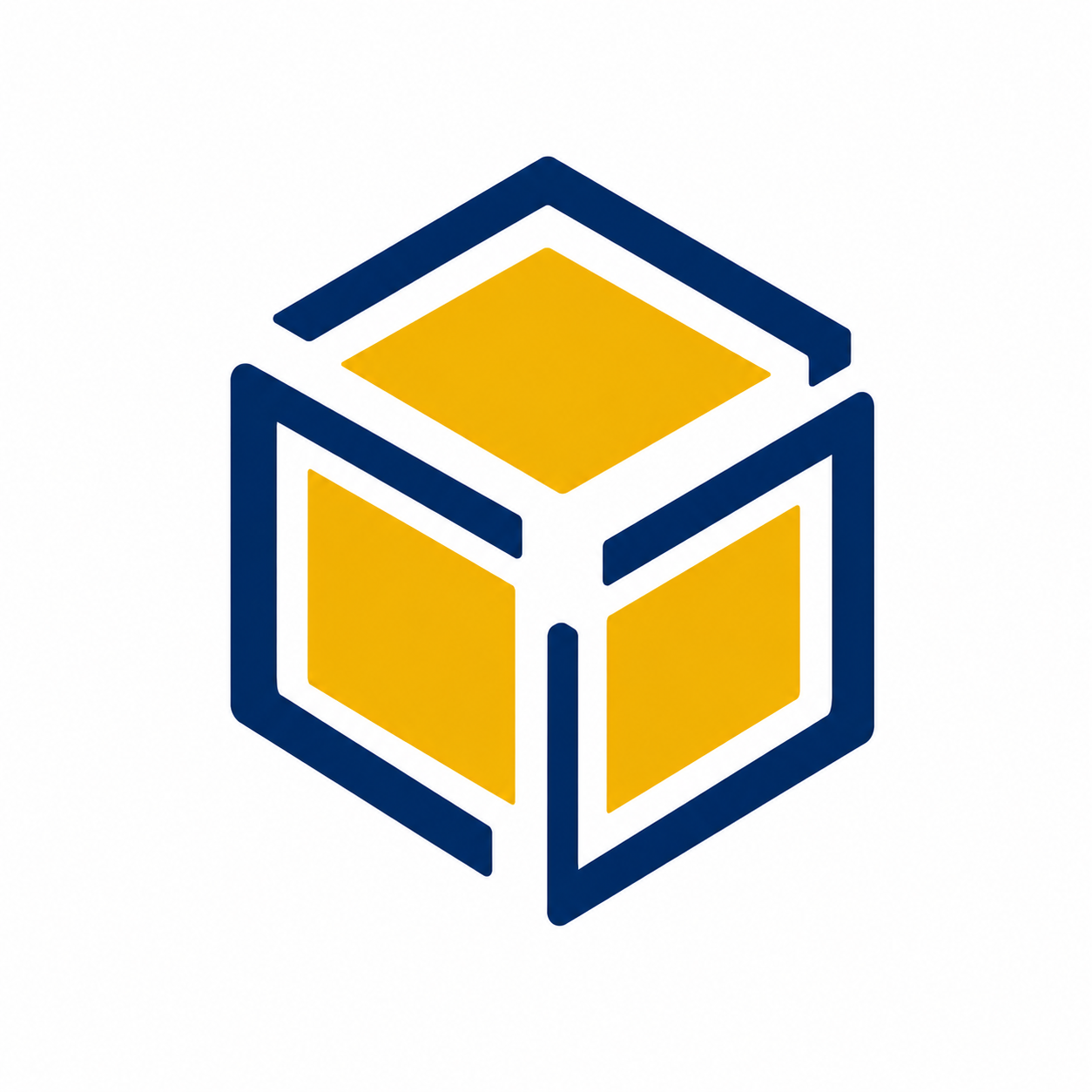}%
   }%
}

\title{
\begin{minipage}{\textwidth}
\centering
\begin{tabular}{c@{\hspace{0.22cm}}l}
\sjtuiicTitleLogo &
\begin{minipage}[c]{0.94\textwidth}
{\fontsize{19.2}{21.2}\selectfont\bfseries
OR-Space: A Full-Lifecycle Workspace Benchmark\\
for Industrial Optimization Agents}
\end{minipage}
\end{tabular}
\end{minipage}
}

\author{%
  Chenyu Zhou\textsuperscript{1,*}\hspace{0.55em}
  Xinyun Lu\textsuperscript{2,*}\hspace{0.55em}
  Jiangyue Zhao\textsuperscript{2,*}\hspace{0.55em}
  Jianghao Lin\textsuperscript{1,\dag}\hspace{0.55em}
  Dongdong Ge\textsuperscript{1}\hspace{0.55em}
  Yinyu Ye\textsuperscript{1,3}\\[2pt]
  {\sjtuiicAffilFont
  \textsuperscript{1}Shanghai Jiao Tong University\quad
  \textsuperscript{2}Shanghai University of Finance and Economics\quad
  \textsuperscript{3}Stanford University}
}

\begin{document}

\begin{abstract}
Large language model (LLM) agents are increasingly used to assist with operations research (OR) modeling, yet existing OR-oriented benchmarks often reduce evaluation to one-shot translation from a self-contained textual problem statement into a mathematical formulation or solver program. Such settings abstract away two defining characteristics of real-world industrial OR workflows: (1) the \textbf{workspace setting}, in which agents must operate within persistent multi-artifact workspaces containing requirements, structured data, code artifacts, solver feedback, and stakeholder interactions; and (2) the \textbf{task setting}, in which agents must support multiple stages of the OR lifecycle rather than only one-shot problem-to-solution generation. To this end, we introduce \textbf{OR-Space}, a full-lifecycle workspace benchmark for evaluating industrial optimization agents across model construction, model revision, and grounded explanation. Each OR-Space instance is represented as a persistent executable workspace containing business documents, structured data, optional code artifacts, solver outputs, and task-specific evaluators distributed across multiple interdependent files. OR-Space defines three lifecycle-oriented task modes: \build, where agents construct solver-ready optimization models from heterogeneous artifacts; \revise, where agents modify existing models under changing requirements or solver feedback while preserving valid prior logic; and \explain, where agents answer grounded questions about solutions, constraints, and business implications using evidence distributed across workspace artifacts. By combining persistent multi-artifact workspaces with lifecycle-oriented task modes, OR-Space evaluates whether agents can perform reliable optimization work beyond end-to-end text generation. We describe the benchmark design, evaluation protocol, and quality-control pipeline, and position OR-Space as a benchmark for studying the reliability, failure modes, and practical readiness of LLM agents in industrial OR workflows.
\end{abstract}

\maketitle

\section{Introduction}
\label{sec:introduction}

Operations research (OR) provides a natural testbed for evaluating the reliability of LLM agents in real-world industrial workflows. In practice, OR work rarely consists of solving a fully specified textbook-style problem in isolation. Instead, OR engineers must interpret business requirements, recover modeling assumptions from documents, extract parameters from structured data files, implement and debug solver programs, revise existing models under evolving requirements, and communicate decisions to non-expert stakeholders. These activities require not only mathematical modeling ability, but also the ability to operate reliably within persistent multi-artifact workspaces.

Existing OR-oriented benchmarks have made substantial progress in evaluating whether LLMs can formulate and solve optimization problems from textual descriptions, covering linear programming, integer programming, mixed-integer programming, and nonlinear programming settings~\cite{huangORLMCustomizableFramework2025,liConstructingIndustrialScale2026}. However, existing OR-oriented LLM evaluations often abstract away key characteristics of real-world industrial OR workflows along two dimensions.
 (1) In terms of the \textbf{workspace setting}, they often compress the relevant business context, data, objective, and constraints into a self-contained textual prompt. This setting is useful for testing end-to-end modeling competence, but it removes the need to locate, connect, and ground information across documents, data files, code artifacts, and solver outputs. (2) In terms of the \textbf{task setting}, existing evaluations often focus on static, single-shot tasks such as generating a formulation or producing solver code from a given problem statement. In practice, OR work is lifecycle-oriented: an engineer builds an initial model, revises it as business rules or data interfaces change, and explains the model and its solution to stakeholders.

Agent benchmarks have demonstrated the importance of evaluating LLM agents in richer environments involving tools, files, execution state, web interfaces, and workplace-style tasks~\cite{jimenezSWEbenchCanLanguage2024,zhouWebArenaRealisticWeb2024,liuAgentBenchEvaluatingLLMs2025,xieOSWorldBenchmarkingMultimodal2024,drouinWorkArenaCapable2024,trivediAppWorldControllable2024,xuTheAgentCompanyBenchmarking2024}. Yet these benchmarks are not designed to evaluate the domain-specific structure of industrial optimization, where correctness depends on mathematical variables, constraints, objectives, data-to-parameter mappings, solver behavior, and faithful explanations grounded in both code and business logic. As a result, current evaluations leave open a central question: can an LLM agent perform reliable OR engineering work across a multi-artifact workspace and across multiple stages of the optimization modeling lifecycle?

We introduce \textbf{OR-Space}, a workspace-grounded benchmark for evaluating LLM agents on industrial optimization tasks. OR-Space is designed around two axes. The first is a \textbf{multi-artifact workspace setting}: each benchmark instance is represented as an executable workspace containing natural language documents, structured data files, optional code artifacts, solver outputs, and automated evaluation interfaces. Rather than receiving a fully serialized problem statement, an agent must recover the relevant optimization problem from distributed artifacts. The second is a \textbf{lifecycle-oriented task setting}: OR-Space evaluates agents across three complementary task modes, \build, \revise, and \explain. In \build, an agent constructs a solver-ready optimization model from business documents and data files. In \revise, the agent modifies an existing workspace in response to new requirements, changed assumptions, or solver feedback while preserving the valid parts of the original model. In \explain, the agent answers grounded questions about the solution, constraints, assumptions, and business implications by reasoning across documents, data, code, and solver output.

This design exposes challenges that are largely hidden in exercise-style OR benchmarks. Agents must ground heterogeneous workspace artifacts into formal optimization structure, maintain cross-artifact consistency under iterative revisions, and produce explanations faithful to both the implemented model and the underlying workspace evidence. These challenges are distinct from, although complementary to, classical formulation and solving ability: an agent may successfully generate optimization code from a clean, fully specified prompt while still failing to operate reliably in realistic industrial OR workspaces.

The goal of OR-Space is to support systematic evaluation of agent reliability in real-world industrial optimization workflows: identifying which stages of the OR lifecycle remain challenging, which failure modes dominate in workspace-grounded environments, and how performance changes when agents must interact with persistent multi-artifact workspaces and execution environments instead of fully specified prompts. OR-Space provides reproducible evaluation across executable model construction, targeted model revision, and grounded explanation. It further enables fine-grained attribution of failures, including execution errors, infeasible formulations, incorrect objectives, missing constraints, incomplete revisions, and unfaithful explanations.

Our contributions are as follows:
\begin{itemize}
    \item We identify two limitations of existing OR-oriented LLM evaluations: self-contained workspace settings and single-shot task settings that underrepresent the multi-artifact and lifecycle-oriented nature of real-world industrial OR workflows.
    \item We introduce \textbf{OR-Space}, a benchmark for industrial optimization agents built around persistent executable workspaces, where requirements, data, code artifacts, solver outputs, and evaluation interfaces are distributed across multiple files and artifacts.
    \item We define three lifecycle-oriented task modes---\build, \revise, and \explain---capturing model construction, model maintenance, and model communication in practical industrial OR workflows.
    \item We provide an evaluation framework for analyzing agent capabilities and failure modes in workspace-grounded OR tasks, including failures in formulation, data grounding, revision consistency, execution, and explanation faithfulness.
\end{itemize}

\section{Related Work}
\label{sec:related}

\paragraph{LLMs for optimization modeling.}
Recent OR benchmarks evaluate LLMs with solver-grounded objectives rather than textual similarity, including NL4Opt~\cite{ramamonjisonNL4OptCompetition2023}, Mamo~\cite{huangLLMsMathematicalModeling2025}, ORLM/IndustryOR~\cite{huangORLMCustomizableFramework2025}, OptiMUS~\cite{ahmaditeshniziOptiMUSScalable2024}, and Chain-of-Experts/ComplexOR~\cite{xiaoChainOfExpertsWhen2024}. Newer work expands this direction through scalable data synthesis, industrial-scale model--data separation, OR question answering, specialized OR training, and agentic modeling or policy-evolution systems~\cite{luOptMATHScalableBidirectional2025,liConstructingIndustrialScale2026,mostajabdavehEvaluatingLLMReasoning2025,zhangORLLMAgentAutomating2025,heEvoOptLLMEvolving2026,huangInvEvolveEvolving2026}. Taken together, these works suggest that robust optimization modeling remains challenging for current LLM systems. However, existing OR modeling benchmarks primarily evaluate isolated, self-contained textbook-style problems. As a result, they often fail to capture failure modes common in industrial optimization workflows, including data extraction, cross-file reasoning, iterative model refinement, and solver-feedback interpretation, which rarely emerge in single-prompt evaluation settings. In contrast to prior NL-to-code optimization benchmarks, OR-Space evaluates persistent optimization reasoning in open-ended workspace environments involving heterogeneous artifacts, execution state, and solver interaction.

\paragraph{Agent evaluation in multi-artifact environments.}
Agent benchmarks such as SWE-bench~\cite{jimenezSWEbenchCanLanguage2024}, WebArena~\cite{zhouWebArenaRealisticWeb2024}, AgentBench~\cite{liuAgentBenchEvaluatingLLMs2025}, GAIA~\cite{mialonGAIABenchmarkGeneral2023}, MLE-bench~\cite{chanMLEbenchEvaluatingMachine2024}, and \taubench~\cite{yaoTauBenchBenchmark2024}, together with survey and usability perspectives on agent evaluation~\cite{zhuEvolutionaryPerspectivesEvaluation2025,liuRealBarrierLLM2025}, demonstrate that realistic environments, tools, execution feedback, and deployment constraints can substantially affect agent behavior and evaluation outcomes. Recent benchmarks extend this paradigm to operating systems, enterprise software, mobile applications, APIs, and simulated organizations~\cite{xieOSWorldBenchmarkingMultimodal2024,drouinWorkArenaCapable2024,trivediAppWorldControllable2024,xuTheAgentCompanyBenchmarking2024}. Related work on agent externalization, protocol standardization, and memory systems further treats reusable skills, protocols, and memory as integral components of the evaluated agent system rather than auxiliary scaffolding~\cite{zhouExternalizationLLMAgents2026,yangSurveyAIAgentProtocols2025,yinMemDecoderEnhancing2026}. OR-Space follows this broader shift toward environment-centric agent evaluation, but specializes it to industrial optimization workflows involving mathematical consistency across optimization formulations, structured data, code, and solver interaction.

\section{OR-Space Benchmark}
\label{sec:dataset}

\begin{figure}[H]
  \centering
  \includegraphics[width=1\textwidth]{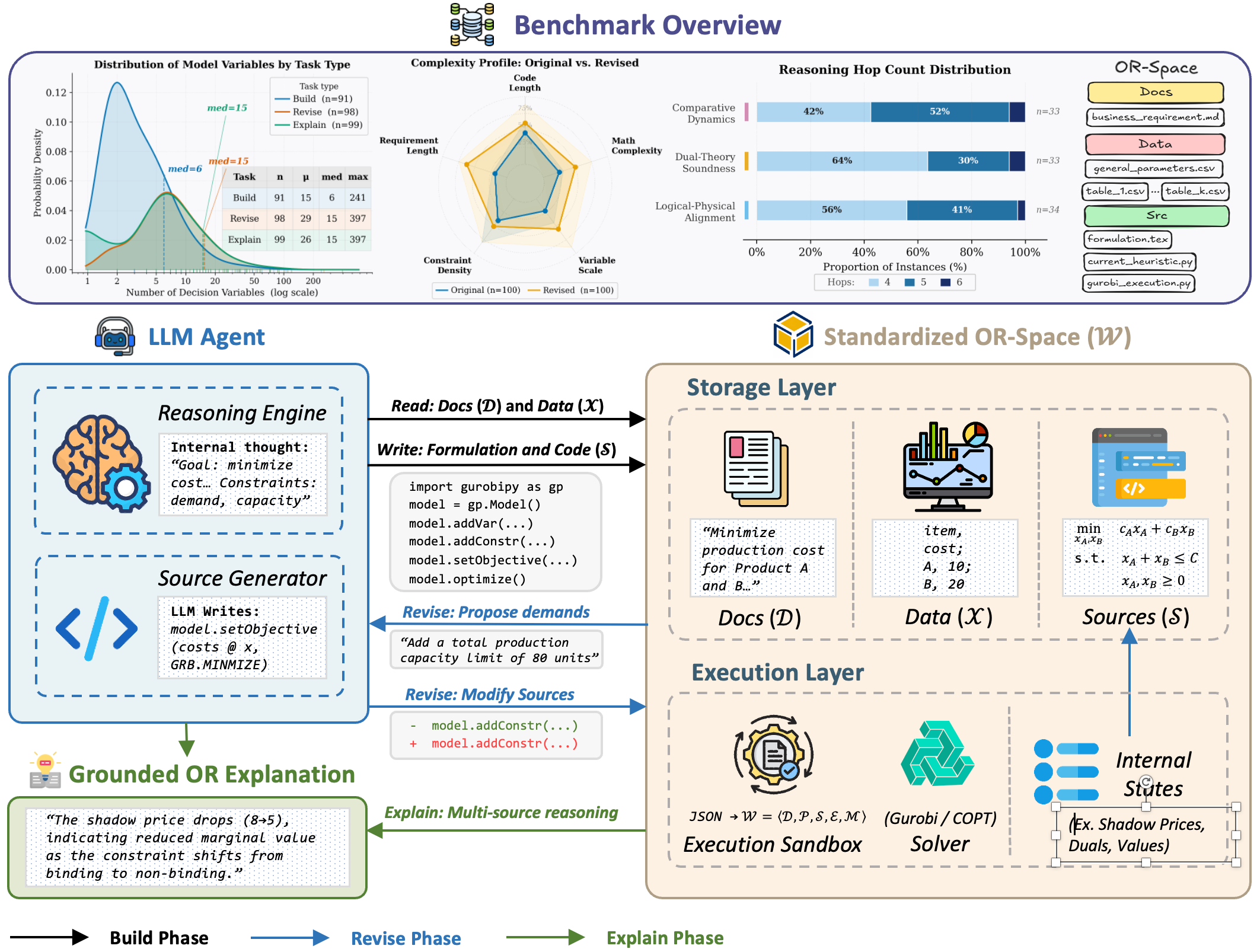} 
  \caption{Overview of the OR-Space benchmark. The figure illustrates the workspace structure, information flow, and artifact dependencies across the \build, \revise, and \explain phases.}
  \label{fig:main}
\end{figure}

In this section, we describe the architecture and evaluation environment of \textbf{OR-Space}, a benchmark designed to evaluate the full lifecycle of industrial optimization modeling. OR-Space evaluates industrial optimization modeling in persistent workspace environments rather than isolated one-shot formulation tasks. The benchmark is organized along two dimensions: an OR workspace composed of documents, parameter files, code artifacts, and solver states, and a lifecycle-oriented task pipeline spanning \build, \revise, and \explain (Figure~\ref{fig:main}).

\subsection{OR Workspace Formalization}
\label{sec:workspace}

We formalize an \emph{OR workspace} as a structured, persistent, executable environment that contains all artifacts relevant to a single OR problem-solving session.

\begin{definition}[OR Workspace]
\label{def:workspace}
An OR workspace is defined as
\begin{equation}
    \workspacedef,
    \label{eq:workspace}
\end{equation}
where $\docset$ denotes document artifacts, $\paramset$ parameter artifacts, $\codeset$ code artifacts, $\envset$ the runtime environment, and $\metricset$ the evaluation metric.
Here, The metric component $\metricset$ is not exposed to the agent as a workspace artifact; it denotes the evaluator attached by the benchmark harness.
\end{definition}

This decomposition follows the model--data separation paradigm used in algebraic optimization systems such as AMPL, Pyomo, and JuMP~\cite{fourerAMPLModelingLanguage2002,hartPyomoModelingSolving2011,dunningJuMPModelingLanguage2017}. OR-Space extends this abstraction by treating separated artifacts, solver execution state, and evaluation signals as part of the interactive benchmark state.

\paragraph{Documents $\docset$.}
Documents are natural-language artifacts describing business requirements, optimization goals, and revision requests. They specify the modeling intent, while numerical values are maintained separately in parameter artifacts, reflecting common industrial workflows.

\paragraph{Parameters $\paramset$.}
Parameter artifacts are structured or semi-structured files, such as CSV or JSON, containing numerical inputs for the optimization model. These files may contain missing values, inconsistent schemas, mixed encodings, or cross-file dependencies requiring cleaning and integration before model construction.

\paragraph{Code $\codeset$ and environment $\envset$.}
Code artifacts may include heuristic scripts, partial model templates, utility functions, or existing optimization programs. Build tasks typically begin from empty scaffolds, while \revise tasks provide executable legacy implementations. The runtime environment is a Docker-based sandbox supporting file I/O, solver execution, stdout/stderr capture, and resource constraints.

\paragraph{Evaluation metric $\metricset$.}
For \build and \revise, evaluation primarily compares the objective value $\agentobj$ produced by the agent against a reference optimum $\refobj$:
\begin{equation}
    \objmetric(\agentobj) = \indicator\!\left[\frac{|\agentobj - \refobj|}{\max\{1, |\refobj|\}} \leq \objtol\right], \qquad \objtol = 0.01.
    \label{eq:eval_obj}
\end{equation}
We additionally report execution and feasibility pass rates. \explain is evaluated using the grounded LLM-as-judge rubric described in Section~\ref{subsec:setup}, including Exact Coverage, Reasoning, Grounding, Answer Quality, and a Hallucination Penalty. Table~\ref{tab:workspace_comparison} compares OR-Space with related OR and agent benchmarks.

\begin{table}[H]
\centering
\caption{Workspace-level comparison of OR-Space with representative OR and agent benchmarks. Components follow $\workspacedef$; \checkmark: present and evaluated, $\sim$: partially present, \texttimes: absent.}
\label{tab:workspace_comparison}
\footnotesize
\setlength{\tabcolsep}{4pt}
\renewcommand{\arraystretch}{1.04}
\begin{tabular}{lccccc}
\toprule
\textbf{Benchmark} & \textbf{Docs} & \textbf{Params} & \textbf{Code} & \textbf{Env.} & \textbf{Metric} \\
 & $\docset$ & $\paramset$ & $\codeset$ & $\envset$ & $\metricset$ \\
\midrule
\sjtuiicgroup{6}{OR modeling benchmarks}
NL4Opt~\cite{ramamonjisonNL4OptCompetition2023}      & $\sim$     & \texttimes & \texttimes & \texttimes & $\sim$     \\
Mamo~\cite{huangLLMsMathematicalModeling2025}        & $\sim$     & \texttimes & \texttimes & \checkmark & \checkmark \\
IndustryOR~\cite{huangORLMCustomizableFramework2025} & $\sim$     & \texttimes & \texttimes & \checkmark & \checkmark \\
ComplexOR~\cite{xiaoChainOfExpertsWhen2024}          & $\sim$     & \texttimes & \texttimes & \checkmark & \checkmark \\
NLP4LP~\cite{ahmaditeshniziOptiMUSScalable2024}      & $\sim$     & \texttimes & \texttimes & \checkmark & \checkmark \\
\addlinespace[0.2em]
\sjtuiicgroup{6}{General agent benchmarks}
SWE-bench~\cite{jimenezSWEbenchCanLanguage2024}      & $\sim$     & \texttimes & \checkmark & \checkmark & \checkmark \\
WebArena~\cite{zhouWebArenaRealisticWeb2024}         & \checkmark & $\sim$     & \texttimes & \checkmark & $\sim$     \\
AgentBench~\cite{liuAgentBenchEvaluatingLLMs2025}    & $\sim$     & $\sim$     & $\sim$     & \checkmark & $\sim$     \\
\midrule
\textbf{OR-Space}                                    & \checkmark & \checkmark & \checkmark & \checkmark & \checkmark \\
\bottomrule
\end{tabular}
\end{table}

\subsection{Lifecycle-Oriented Tasks}
\label{sec:task_modes}

OR-Space organizes industrial optimization workflows into three sequential task modes corresponding to model construction, structural modification, and solver-grounded interpretation.

\taskwithsymbol{Build}{\buildtask} In the \build setting, the agent receives requirement documents in $\docset$, parameter spreadsheets in $\paramset$, and an empty code scaffold in $\codeset$. The goal is to construct a complete optimization model by parsing natural-language requirements, aligning data schemas, and implementing variables, constraints, and objectives within a unified \pulpmodel interface. The evaluation environment $\envset$ then attaches the configured reference solver backend (e.g., Gurobi, with cross-solver validation on COPT or HiGHS) to execute runtime verification.

\taskwithsymbol{Revise}{\revisetask} In the \revise setting, the agent modifies an existing optimization workflow under updated business requirements. The workspace includes revised documents and data schemas together with a legacy implementation of the baseline problem. The agent must update the existing model while preserving unaffected logic, supporting operations such as variable insertion, constraint modification, and multi-period extensions. OR-Space includes three variants: \revisecode (code only), \revisemodel (formulation only), and \reviseall (code plus formulation).

\taskwithsymbol{Explain}{\explaintask} In the \explain setting, the agent receives the full workspace together with solver outputs from $\envset$, including logs, feasibility status, runtime statistics, dual variables, and constraint slacks. The task is to generate short factual reports explaining bottlenecks, sensitivities, or allocation decisions based on the executed optimization model. Unlike \build and \revise, which are solver-scored, \explain evaluates grounded natural-language reasoning tied directly to solver states and optimization behavior.

\subsection{Benchmark Construction Pipeline}
\label{subsec:generation}

We next describe how the three lifecycle settings are constructed from the underlying IndustryOR problems while preserving a shared mathematical structure across task instances. OR-Space extends the 100 base optimization problems from the IndustryOR benchmark~\cite{huangORLMCustomizableFramework2025} into multi-artifact \build, \revise, and \explain task instances, yielding 100 instances per setting and 300 instances in total. Each instance is generated through a two-stage pipeline: we first construct a clean specification closely aligned with the ground-truth mathematical model, and then rewrite it into a realistic business-facing workspace containing domain terminology, conversational redundancies, inconsistent schema references, and noisy organizational language.

\taskheading{Build}
For \build, each instance is first converted into the unified OR-Space workspace representation
\[
\workspacedef,
\]
where problem descriptions, parameter tables, executable code, runtime environments, and evaluation records are explicitly separated into heterogeneous artifacts. We additionally construct verified algebraic formulations and executable reference implementations as hidden oracle records for evaluation. The agent only observes the workspace-facing artifacts and must recover the optimization model through schema alignment, parameter grounding, and executable model construction.

\taskheading{Revise}
For \revise, we synthesize programmatic requirements evolution over existing workspace instances. The generation pipeline introduces coupled structural updates involving simultaneous insertion, deletion, and modification of variables and constraints, such as adding and removing cities in routing problems or extending models to multi-period settings. Many revisions introduce multi-variable constraint coupling, where newly added business rules alter existing constraint relations rather than acting independently. To preserve correctness, revised instances undergo double-build execution checks during synthesis to ensure that new constraints do not unintentionally invalidate unaffected model components or collapse feasibility regions.

\taskheading{Explain}
For \explain, we construct grounded reasoning tasks directly from verified solver executions. The pipeline extracts solver-derived signals including dual variables, binding constraints, slack values, Big-M tightness conditions, and LP-relaxation bounds, and then generates business-oriented analytical questions requiring multi-hop reasoning across workspace artifacts and optimization states. These tasks are explicitly coupled with foundational OR concepts such as duality theory, constraint activity, sensitivity analysis, and relaxation behavior, ensuring that explanations remain grounded in the executed mathematical model rather than surface-level textual patterns.

To improve data quality, generated tasks are additionally reviewed by researchers with operations research experience. The review process checks consistency across business descriptions, parameter schemas, solver execution states, and explanation rubrics, helping reduce inconsistencies across workspace artifacts and solver-grounded evaluation signals.

\section{Experiments}
\label{sec:experiments}

\subsection{Experiment Setup}
\label{subsec:setup}

\paragraph{Evaluation goal and default track.}
We evaluate whether LLM agents can operate over the full OR-Space lifecycle,
covering model construction (\build), requirement revision (\revise), and
solver-grounded explanation (\explain). Unless otherwise specified, all
headline results use the filesystem workspace interface, the
\revisecode setting, and Gurobi~12.0.1~\cite{gurobi} as the default
solver backend and ground-truth oracle. Controlled variants are introduced in
the result subsections.

\paragraph{Agents and runtime.}
We evaluate 20 models spanning closed-source frontier models, small API-based
models, and open-source models; the full list
appears in Table~\ref{tab:main_results}. The model set includes both standard
and explicit reasoning or thinking variants where available. Every agent runs
in an isolated workspace with read--write access restricted to its own
\texttt{docs/data/src/} subtree and no network access. API-based models use
temperature $\le 0.1$; code-generation calls use $32{,}768$ completion tokens
except GPT-4o, capped at $16{,}384$. \build and \revise submissions execute in
a fresh Python interpreter under a 120\,s wall-clock limit; the harness records
solver status and the objective value.

\begin{figure}[H]
  \centering
  \includegraphics[width=1\textwidth]{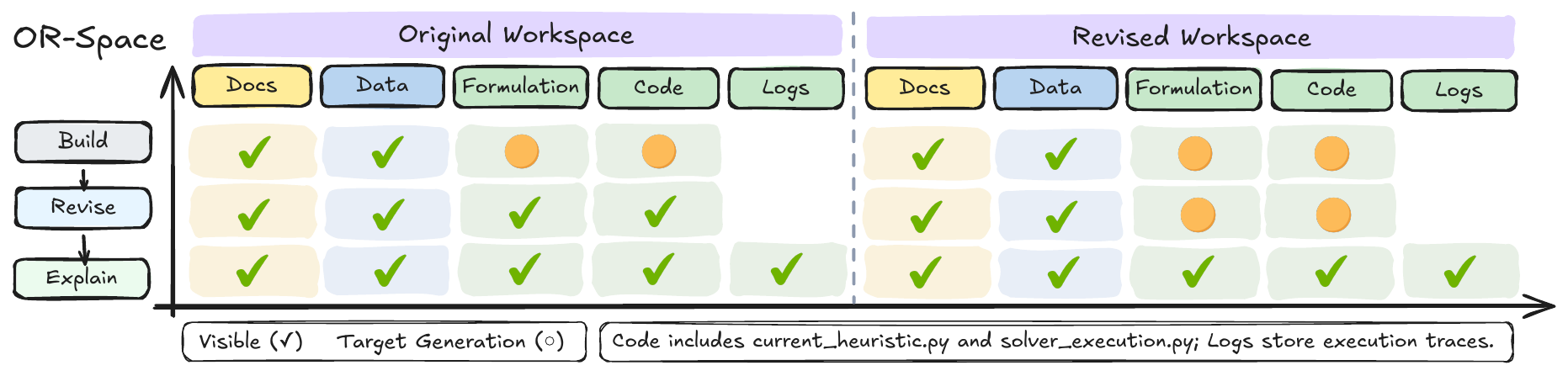}
  \caption{Task-artifact visibility across the OR-Space workflow. The matrix defines the experimental setup and information accessibility for the \build, \revise, and \explain phases.}
  \label{fig:task_visibility}
\end{figure}

\paragraph{Task Visibility.}
The three tasks defined in Section~\ref{sec:task_modes} differ in the visibility of workspace artifacts provided to the agent (Figure~\ref{fig:task_visibility}). \build exposes business documents and data without any completed solver model, requiring the agent to construct the optimization logic from scratch. \revise exposes the original workspace and revised requirements, including legacy heuristic code as contextual information, while withholding the revised reference model. \explain provides both the original and revised workspaces together with solver records, requiring the agent to ground its responses in the available documents, data, code, and solver outputs.

\paragraph{Evaluation metrics.}
\build and \revise are solver-scored. We adopt the objective-matching metric
from Section~\ref{sec:dataset}: a submission is counted as correct
($\objmetric = 1$) iff the script executes without error, returns
an \optimalstatus status, and achieves relative objective error at most
$\objtol = 10^{-2}$ with respect to the Gurobi reference solution $\refobj$,
i.e., $|\agentobj - \refobj| / \max(1, |\refobj|) \le \objtol$. All submissions
expose optimization models via a unified \pulpmodel interface,
while the evaluation harness attaches the solver backend at runtime,
decoupling modelling correctness from solver-specific execution. We
additionally log failure modes including \wrongvalue,
\runtimeerror, \emptyoutput, and \apiexception for
diagnostic analysis.

\explain is judge-assisted but grounded. Each instance ships with a
ground-truth checklist combining
(i) \texttt{exact\_match} items (variable names, constraint names, CSV
columns, numeric values) scored programmatically after text normalisation,
and (ii) \texttt{llm\_boolean\_judgment} items scored by an independent judge
model against a criterion-specific rubric. We report the five-dimensional
Rubric mean score ($\expmetric \in [0, 100]$): Exact Coverage
(35), Reasoning (35), Grounding (20), Answer Quality (10), minus a
Hallucination Penalty (up to $-12$). LLM-as-judge is therefore restricted to
short, workspace-grounded explanations rather than used for the solver-scored
construction tasks.

\begin{table}[!htbp]
\centering
\ifarxivversion
\captionsetup{font=small,skip=2pt}
\caption{Main OR-Space results on 100 instances. \build and \revise report
Pass@1 against the Gurobi oracle; \explain reports the five-dimensional Rubric
mean. \revise is \revisecode. Higher is better; best values are highlighted in \textbf{bold}, and second-best values are \secondval{underlined} within each model category.}
\fi
\label{tab:main_results}
\ifarxivversion
\footnotesize
\setlength{\tabcolsep}{3.2pt}
\renewcommand{\arraystretch}{0.92}
\fi
\begin{tabular}{lcccc}
\toprule
\textbf{Model} & \taskbold{Build} & \taskbold{Revise (code)} & \taskbold{Explain} & $\passdelta$ \\
 & Pass@1 (\%, $\uparrow$) & Pass@1 (\%, $\uparrow$) & Rubric (0--100, $\uparrow$) & (pp, $\uparrow$) \\
\midrule
\sjtuiicgroup{5}{Closed-source Models}
\texttt{gemini-3.1-pro}            & \bestval{72.0} & \bestval{81.0} & 73.00 & $+9$ \\
\texttt{gemini-3-flash}            & \secondval{68.0} & 45.0 & 53.24 & $-23$ \\
\texttt{claude-opus-4-6}           & 62.0 & \secondval{80.0} & 80.20 & $+18$ \\
\texttt{gpt-5.4}                   & 59.0 & 79.0 & \bestval{86.52} & $+20$ \\
\texttt{gpt-5-mini}                & 58.0 & 77.0 & \secondval{82.94} & $+19$ \\
\texttt{claude-sonnet-4.5}         & 53.0 & 78.0 & 77.58 & $+25$ \\
\texttt{claude-sonnet-4.5-thinking}& 53.0 & 76.0 & 79.63 & $+23$ \\
\texttt{gpt-5.1}                   & 53.0 & 71.0 & 76.16 & $+18$ \\
\texttt{gemini-2.5-flash}          & 53.0 & 66.0 & 13.79 & $+13$ \\
\texttt{gemini-2.5-pro}            & 47.0 & 69.0 & 32.22 & $+22$ \\
\texttt{gpt-4o}                    & 25.0 & 36.0 & 59.36 & $+11$ \\
\midrule
\sjtuiicgroup{5}{Open-source Models}
\texttt{deepseek-v4-pro}           & \bestval{67.0} & \bestval{71.0} & 68.87 & $+4$ \\
\texttt{deepseek-r1-0528}          & \secondval{55.0} & \secondval{65.0} & 44.40 & $+10$ \\
\texttt{deepseek-v4-flash}         & 54.0 & 63.0 & \secondval{77.77} & $+9$ \\
\texttt{qwen3-max}                 & 49.0 & 60.0 & 74.66 & $+11$ \\
\texttt{qwen3-32b}                 & 32.0 & 20.0 & 61.32 & $-12$ \\
\texttt{qwen3-32b-thinking}        & 28.0 & 32.0 & 61.28 & $+4$ \\
\texttt{qwen3.5-27b}               & 25.0 & 48.0 & \bestval{77.88} & $+23$ \\
\texttt{qwen3-8b}                  & 22.0 & 12.0 & 55.19 & $-10$ \\
\texttt{qwen3-14b}                 & 20.0 & 19.0 & 61.03 & $-1$ \\
\bottomrule
\end{tabular}
\end{table}

\subsection{Main Results Across \taskbold{Build}, \taskbold{Revise}, and \taskbold{Explain}}
\label{subsec:main_results_discussion}

We evaluate agents on the three lifecycle tasks under the Gurobi
track. Table~\ref{tab:main_results} reports \build, \revisecode, and
\explain performance for all 20 models. From these results, several patterns emerge:

\textbf{Firstly,
workspace-grounded model construction remains difficult even for
frontier agents}. The strongest \build result is $72.0\%$
(\texttt{gemini-3.1-pro}), and most frontier models remain in the
$53$--$68\%$ range. These instances reuse IndustryOR-derived mathematical
topologies, but the \build task no longer presents the problem as a
self-contained formulation prompt. Instead, the agent must recover the data
schema, business semantics, and model structure jointly from \texttt{docs/},
\texttt{data/}, and an empty \texttt{src/}. This is the central workspace
difficulty that the error analysis later makes explicit. 

\textbf{Secondly,
\taskbold{Revise} exposes whether a model can use legacy code as
signal rather than noise}. Strong models often benefit from the additional
heuristic context: \texttt{gemini-3.1-pro} improves from $72\%$ to
$81\%$, and \texttt{gpt-5.4} from $59\%$ to $79\%$. The same setting can
hurt weaker or flash-tier models, however. \texttt{gemini-3-flash}
drops by $23$\,pp and \texttt{qwen3-32b} by $12$\,pp, suggesting that a
working heuristic is useful only when the agent can separate reusable
mathematical structure from implementation details that should not be copied.

\textbf{Additionally, \taskbold{Explain} measures solution-grounded understanding rather than another version of model construction.} \texttt{gpt-5.4} obtains the highest \explain score ($86.52$), while some models that build runnable models explain them poorly: \texttt{gemini-2.5-pro} solves $69\%$ of \revise instances but scores $32.2$ on \explain, and \texttt{gemini-2.5-flash} reaches $66\%$ \revise but only $13.79$ on \explain. This highlights \explain’s ability to capture reasoning over distributed workspace artifacts, revealing capabilities that \build or \revise alone cannot measure. As noted in Section~\ref{subsec:filesystem_vs_flat}, \explain’s performance under Filesystem vs Flat-prompt differs from \build/\revise, emphasizing its unique role in assessing workspace-grounded reasoning.

\textbf{The correlations reinforce this separation between tasks:}
\build and \revise are strongly aligned ($r_{\text{B,R}}=0.82$), whereas \explain
is weakly correlated with both ($r_{\text{B,E}}=0.16$,
$r_{\text{R,E}}=0.28$). Evaluating only solver-correct model construction would therefore miss whether an agent can faithfully explain the optimization behavior it produces across the full workspace.


\subsection{Workspace Setting Matters: Filesystem vs.\ Flat Prompt}
\label{subsec:filesystem_vs_flat}

To isolate the effect of the workspace interface, OR-Space’s default setting exposes documents, data, source files, and solver records as separate artifacts; the \emph{Flat} variant serializes the same workspace into a single JSON-style prompt (including the directory tree and the full contents of every Markdown, CSV, and Python file, concatenated) without filesystem tools. Table~\ref{tab:fs_vs_flat} compares these two interfaces on a controlled six-model subset.

\begin{table}[!htbp]
\centering
\caption{Filesystem (\fsvariant) vs.\ Flat-prompt (\flatvariant) evaluation on
the controlled 6-model subset. \textsc{Build} and \textsc{Revise} are Pass@1 (\%); \textsc{Explain} is the
Rubric mean score. \textsc{Revise} uses the same \revisecode setting as the
main table, where the agent sees the original heuristic source plus the revised
business requirement, but not the revised reference code. Best values are highlighted in \textbf{bold}, and second-best values are \secondval{underlined}.}
\label{tab:fs_vs_flat}
\footnotesize
\setlength{\tabcolsep}{4pt}
\renewcommand{\arraystretch}{1.04}
\begin{tabular}{lcccccc}
\toprule
\textbf{Model}
& \textbf{\textsc{Build}} & \textbf{\textsc{Build}}
& \textbf{\revisecode} & \textbf{\revisecode}
& \textbf{\textsc{Explain}} & \textbf{\textsc{Explain}} \\
& \fsvariant ($\uparrow$) & \flatvariant ($\uparrow$)
& \fsvariant ($\uparrow$) & \flatvariant ($\uparrow$)
& \fsvariant ($\uparrow$) & \flatvariant ($\uparrow$) \\
\midrule
\texttt{gemini-3.1-pro}  & \bestval{72} & \bestval{72} & \bestval{81} & \bestval{94} & 73.00 & 72.55 \\
\texttt{deepseek-v4-pro}         & \secondval{67} & \secondval{70} & 71 & 85 & 68.87 & 68.76 \\
\texttt{gpt-5.4}                 & 59 & 66 & \secondval{79} & \secondval{89} & \bestval{86.52} & \bestval{86.17} \\
\texttt{claude-sonnet-4.5}       & 53 & 62 & 78 & 85 & \secondval{77.58} & 75.68 \\
\texttt{qwen3-max}               & 49 & 60 & 60 & 75 & 74.66 & \secondval{77.19} \\
\texttt{gpt-4o}                  & 25 & 44 & 36 & 50 & 59.36 & 57.55 \\
\midrule
\textit{Average \fsflatdelta}
& $-8.2$ & & $-12.2$ & & $+0.3$ & \\
\bottomrule
\end{tabular}
\end{table}

The comparison shows a task-dependent interface effect. 
\textbf{Specifically, In the code-producing
tasks, the filesystem interface lowers performance mainly when agents must
produce executable code}. \build loses $8.2$\,pp on average under the
filesystem interface, with a $19$\,pp gap for \texttt{gpt-4o} and an $11$\,pp
gap for \texttt{qwen3-max}. Similarly, The penalty is even larger for
\revisecode ($-12.2$\,pp), despite the presence of the original
heuristic. This indicates that file discovery, schema inspection, and
path-correct code generation are not cosmetic parts of the interface; they are
substantive sources of difficulty in workspace-based OR modelling. 

\textbf{By contrast, flattening helps little for \explain because the evidence-alignment problem remains.} On average, \explain changes by $+0.3$,pp. This occurs because \explain inherently relies on cross-file reasoning similar to \revise, so the structured filesystem helps clarify relationships across documents, data, and code. Flattening removes the directory structure but does not eliminate the need to integrate these artifacts, and the clearer organization in the filesystem can offset the difficulty of cross-file reasoning, explaining why \explain does not suffer a performance drop like \build or \revise.

\subsection{Revision Context Is Double-Edged}
\label{subsec:heuristic_context}

To understand the impact of legacy code and formal formulations on model performance, this subsection compares three \revise settings: \revisecode, \reviseall, and \revisemodel.
Relative to \build, these settings add the original heuristic source or formal formulations while keeping the instance family and objective oracle fixed. 
The $\passdelta$ column in Table~\ref{tab:main_results} provides a measure of the net effect of legacy-code context for \revisecode, and similar comparisons can be made for the other Revise variants. 

The revision context raises a natural question: can a formal formulation replace
the heuristic, or does it provide a different kind of signal?
Table~\ref{tab:revise_context_ablation_corrected} reports macro-averaged
Pass@1 separately for all models, high-capability models, and lightweight
models, with the last two columns measuring the gain or loss relative to
\revisecode.

\begin{table}[!htbp]
\centering
\caption{Solver-level \revise context ablation: corrected average performance for high-capability and lightweight models. Best values are highlighted in \textbf{bold}, and second-best values are \secondval{underlined} within each model group.}
\label{tab:revise_context_ablation_corrected}
\footnotesize
\setlength{\tabcolsep}{4pt}
\renewcommand{\arraystretch}{1.04}
\begin{tabular}{lccccc}
\toprule
\textbf{Solver} &
\textbf{\revisecode ($\uparrow$)} &
\textbf{\reviseall ($\uparrow$)} &
\textbf{\revisemodel ($\uparrow$)} &
\textbf{All $-$ Code ($\uparrow$)} &
\textbf{Model $-$ Code ($\uparrow$)} \\
\midrule
\sjtuiicgroup{6}{All models}
Gurobi & \secondval{57.4} & \bestval{64.8} & \bestval{59.0} & +7.4 & +1.6 \\
COPT   & 49.1 & 52.4 & 46.6 & +3.2 & $-2.5$ \\
PuLP   & \bestval{59.0} & \secondval{62.1} & \secondval{53.5} & +3.1 & $-5.5$ \\
HiGHS  & 52.1 & 48.6 & 50.6 & $-3.5$ & $-1.5$ \\
\midrule
\sjtuiicgroup{6}{High-capability models}
Gurobi & \secondval{70.6} & \bestval{80.6} & \bestval{76.5} & +10.0 & +5.9 \\
COPT   & 69.8 & 75.7 & 67.9 & +5.9 & $-1.9$ \\
PuLP   & \bestval{73.6} & \secondval{77.0} & \secondval{68.7} & +3.4 & $-4.9$ \\
HiGHS  & 67.1 & 62.7 & 66.2 & $-4.4$ & $-0.9$ \\
\midrule
\sjtuiicgroup{6}{Lightweight models}
Gurobi & \secondval{44.2} & \bestval{49.0} & \bestval{41.6} & +4.8 & $-2.6$ \\
COPT   & 28.4 & 29.0 & 25.4 & +0.6 & $-3.0$ \\
PuLP   & \bestval{44.4} & \secondval{47.3} & \secondval{38.4} & +2.9 & $-6.0$ \\
HiGHS  & 37.2 & 34.5 & 35.1 & $-2.7$ & $-2.1$ \\
\bottomrule
\end{tabular}
\end{table}

\textbf{Adding formal formulations on top of executable heuristic code is generally beneficial, although the magnitude of the benefit depends on both the solver backend and model capability.}
Compared with \revisecode, \reviseall improves average Pass@1 on Gurobi, COPT, and PuLP by $+7.4$, $+3.2$, and $+3.1$\,pp, respectively, while reducing performance on HiGHS by $3.5$\,pp. This gain is more pronounced for high-capability models, which benefit substantially from the added formulation context on Gurobi, COPT, and PuLP, whereas lightweight models exhibit smaller but still positive improvements on the same solvers. These results suggest that stronger models can leverage formal formulations to recover mathematical structure that is only partially exposed by the heuristic implementation, rather than merely imitating existing code patterns.

\textbf{However, formulation-only context is substantially more brittle than code-based context.} In the all-model aggregate, \revisemodel outperforms \revisecode only on Gurobi ($+1.6$\,pp) and underperforms it on COPT, PuLP, and HiGHS. This gap is
especially clear for lightweight models, where \revisemodel is worse than \revisecode under all four solvers. These results indicate that formal formulations alone do not reliably replace executable heuristic code.

Taken together, these results suggest that \textbf{formal formulations and executable code provide complementary, rather than interchangeable, context}.
Formulations expose the intended mathematical structure, whereas executable
heuristics expose indexing, data loading, and implementation conventions.
Current agents need both kinds of signal, and their ability to exploit each
signal depends on model capability and solver backend.

\begin{figure}[!htbp]
  \centering
  \includegraphics[width=0.9\textwidth]{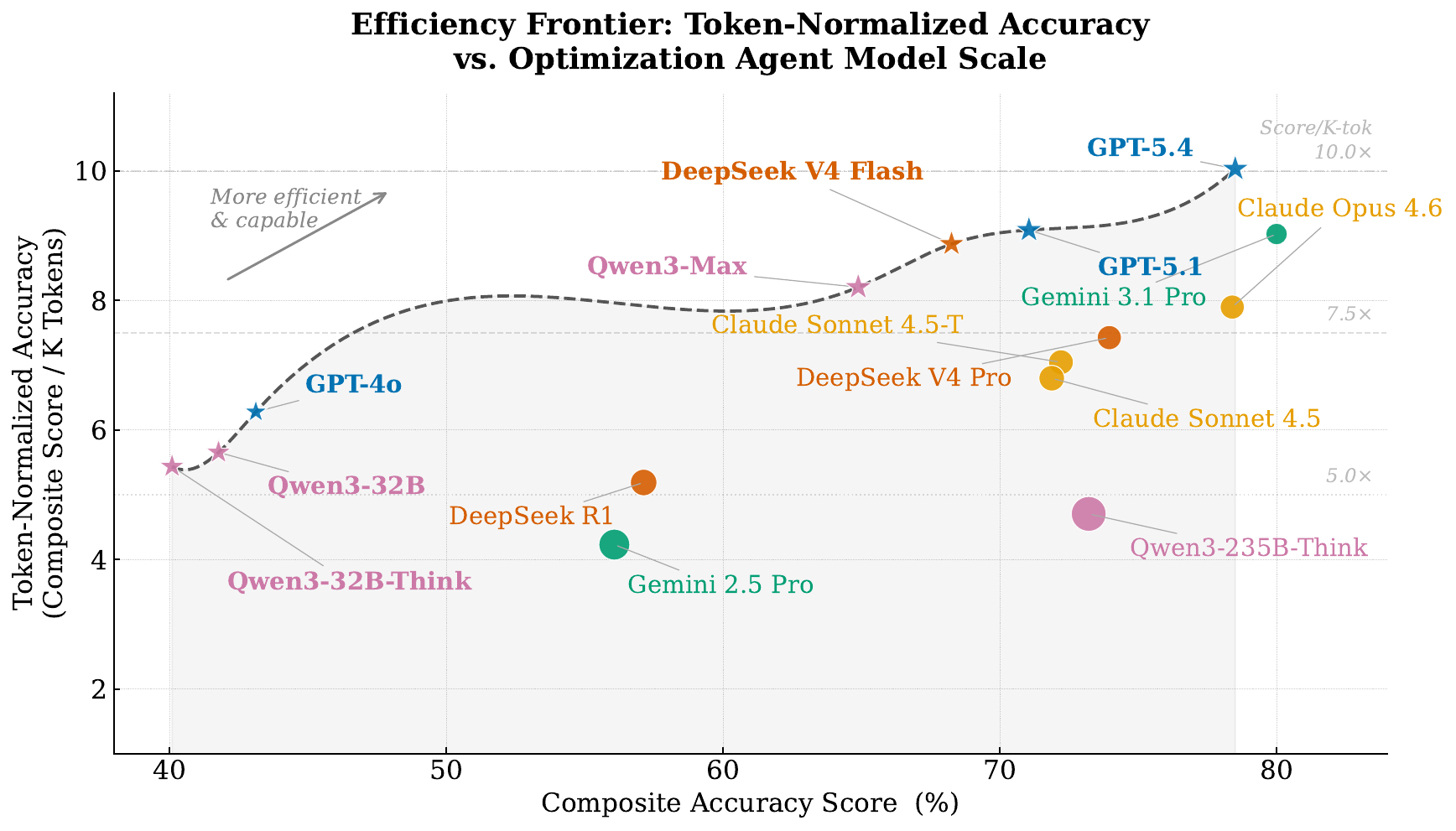}
  \caption{The Efficiency Frontier of Model Scaling. Stars denote Pareto-optimal configurations, representing the upper bound of efficiency. Marker size scales with the average number of tokens consumed per instance. }
  \label{fig:token}
\end{figure}

Figure~\ref{fig:token} highlights two model-level trends from the individual model analysis. \textbf{Firstly, legacy code is valuable only for models that can abstract from it.}
Stronger agents are often context-amplifying:
\texttt{claude-sonnet-4.5} ($+25$), \texttt{gpt-5.4} ($+20$),
\texttt{claude-opus-4-6} ($+18$), and \texttt{gemini-2.5-pro} ($+22$) appear
to recover useful structure from the heuristic, such as variable granularity,
join keys, and period indexing. \texttt{deepseek-v4-pro} and
\texttt{qwen3-32b-thinking} are closer to context-neutral (both $+4$). In
contrast, \texttt{gemini-3-flash} ($-23$) and \texttt{qwen3-32b} ($-12$) are
context-poisoned, frequently carrying over non-solver imports or dangling
heuristic variable names.

\textbf{More broadly, the same additional context
has opposite marginal utility across models}. To make this effect comparable
across prompt lengths, we compute the per-token marginal accuracy gain
$\text{CMU} \coloneqq (\text{Pass}_{\text{\revise}} - \text{Pass}_{\text{\build}})
/(\text{prompt}_{\text{\revise}} - \text{prompt}_{\text{\build}})\times10^3$.
In this view, \texttt{gpt-5.4} reaches $+11.9$ and
\texttt{claude-sonnet-4.5} reaches $+11.3$ accuracy points per thousand
additional tokens, whereas \texttt{gemini-3-flash} and
\texttt{qwen3-32b} fall to $-11.9$ and $-7.0$. The same information source can
therefore act as signal for one model and noise for another.

\subsection{Cross-Solver Robustness}
\label{subsec:solver_robustness}

To assess whether the findings depend on a single solver backend,
Table~\ref{tab:solver_comparison} aggregates the complete four-backend results
from Appendix~\ref{app:full_solver_backend_results}, using macro-averages over
the completed model rows.

\begin{table}[!htbp]
\centering
\caption{Solver comparison on OR-Space by model group. Entries are macro-averages over the completed model-level results for each solver track in Appendix Table~\ref{tab:appendix_solver_full}. \objavg averages \build and the three \revise variants; Range is the max--min spread across all five task columns. Best values are highlighted in \textbf{bold}, and second-best values are \secondval{underlined} within each model group.}
\label{tab:solver_comparison}
\footnotesize
\setlength{\tabcolsep}{3pt}
\renewcommand{\arraystretch}{1.04}
\begin{tabular}{lccccccc}
\toprule
\textbf{Solver} & \textbf{\build ($\uparrow$)} & \textbf{\textsc{R-model} ($\uparrow$)} & \textbf{\textsc{R-code} ($\uparrow$)} & \textbf{\textsc{R-all} ($\uparrow$)} & \textbf{\explain ($\uparrow$)} & \textbf{\objavg ($\uparrow$)} & \textbf{Range ($\downarrow$)} \\
\midrule
\sjtuiicgroup{8}{All models}
\texttt{Gurobi} & \bestval{47.8} & \bestval{59.0} & \secondval{57.4} & \bestval{64.8} & \bestval{64.9} & \bestval{57.2} & 17.1 \\
\texttt{COPT}   & 38.3 & 46.6 & 49.1 & 52.4 & \secondval{57.1} & 46.6 & 18.8 \\
\texttt{PuLP}   & \secondval{47.2} & \secondval{53.5} & \bestval{59.0} & \secondval{62.1} & 56.1 & \secondval{55.5} & \secondval{14.9} \\
\texttt{HiGHS}  & 45.4 & 50.6 & 52.1 & 48.6 & 55.2 & 49.2 & \bestval{9.9} \\
\midrule
\sjtuiicgroup{8}{High-capability models}
\texttt{Gurobi} & \bestval{59.1} & \bestval{76.5} & \secondval{70.6} & \bestval{80.6} & \bestval{71.4} & \bestval{71.7} & 21.5 \\
\texttt{COPT}   & 56.5 & 67.9 & 69.8 & 75.7 & \secondval{65.2} & 67.5 & \secondval{19.2} \\
\texttt{PuLP}   & \secondval{57.1} & \secondval{68.7} & \bestval{73.6} & \secondval{77.0} & 63.7 & \secondval{69.1} & 19.9 \\
\texttt{HiGHS}  & 55.4 & 66.2 & 67.1 & 62.7 & 62.1 & 62.8 & \bestval{11.7} \\
\midrule
\sjtuiicgroup{8}{Lightweight models}
\texttt{Gurobi} & \secondval{36.4} & \bestval{41.6} & \secondval{44.2} & \bestval{49.0} & \bestval{58.3} & \bestval{42.8} & 21.9 \\
\texttt{COPT}   & 20.1 & 25.4 & 28.4 & 29.0 & \secondval{49.0} & 25.7 & 28.9 \\
\texttt{PuLP}   & \bestval{37.3} & \secondval{38.4} & \bestval{44.4} & \secondval{47.3} & 48.5 & \secondval{41.8} & \bestval{11.2} \\
\texttt{HiGHS}  & 35.3 & 35.1 & 37.2 & 34.5 & 48.3 & 35.5 & \secondval{13.8} \\
\bottomrule
\end{tabular}
\end{table}

\textbf{Solver choice changes absolute scores and task profiles, even when the broad ranking is stable.} 
Gurobi has the highest objective-task average (57.2) and the highest \explain average (64.9). 
PuLP is close on objective tasks and gives the best average \revisecode score, likely because the evaluation uses PuLP’s own heuristic code, giving it a more aligned context for code-guided model revision. 
COPT, meanwhile, shows performance comparable to Gurobi for high-capability models, indicating that well-trained models can effectively utilize COPT's API for code-guided revision. However, for lightweight models, performance on COPT is lower than on Gurobi, likely due to the higher exposure of Gurobi in the pretraining corpora; this reflects differences in model familiarity with solver-specific code rather than any difference in the solvers' intrinsic mathematical capabilities. 
HiGHS has the smallest cross-task range (9.9), suggesting a more even but lower profile across tasks.

\textbf{Regarding ranking consistency}, the complete model-level table shows that Gurobi and COPT rankings are broadly aligned ($\rankcorr = 0.73$), although absolute deltas can be large for individual agents. Thus, solver choice is not merely an implementation detail: a single backend can mix modelling ability with solver-interface sensitivity.

\subsection{Failure Analysis: What Workspace Evaluation Reveals}
\label{sec:failure_analysis}
\label{subsec:failure_modes}

\begin{figure}[!htbp]
  \centering
  \includegraphics[width=1\textwidth]{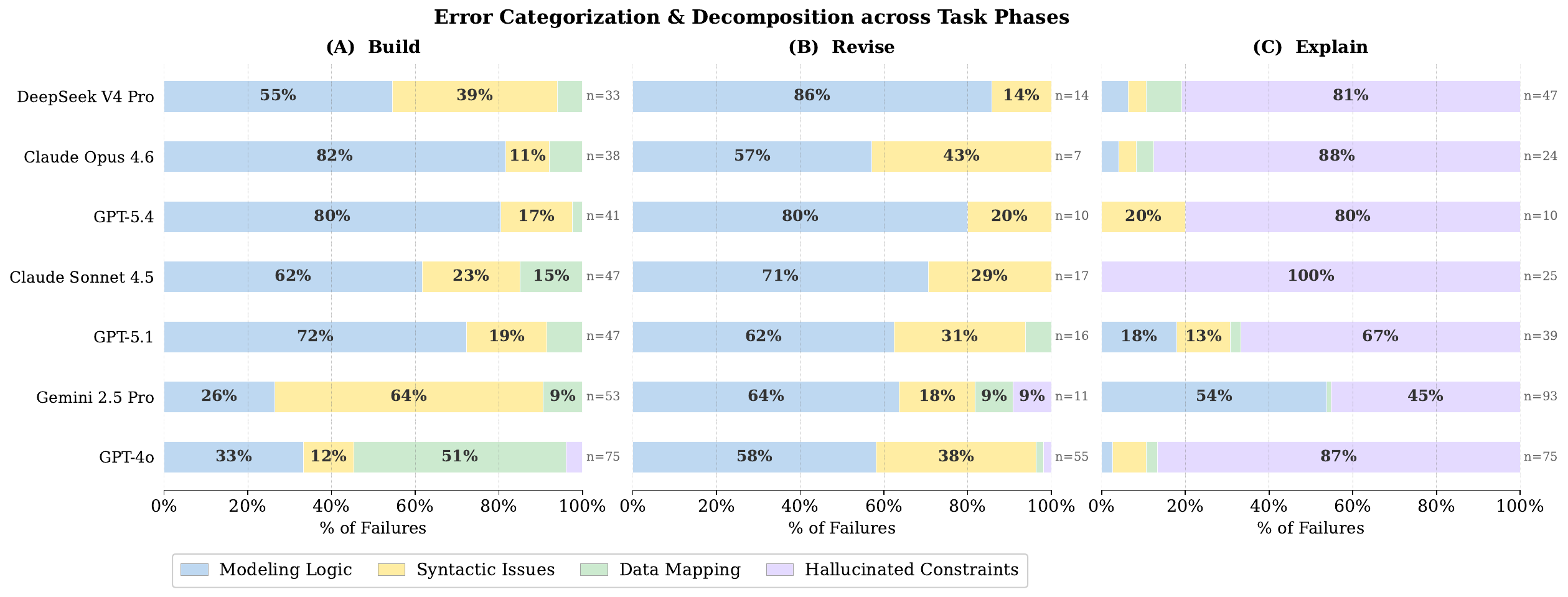}
  \caption{Error Categorization and Decomposition across \build, \revise, and \explain Tasks.}
  \label{fig:error_analysis}
\end{figure}

Failure analysis identifies what creates the gap between clean-prompt OR
modelling and workspace-grounded OR agents.
Figure~\ref{fig:error_analysis} summarizes semantic failure categories for
representative models across \build, \revise, and \explain. Each horizontal bar shows the distribution of categorized failures for one model
within a task phase, and the right-side $n$ reports the number of failed
instances being categorized. The categories separate modelling-logic mistakes, syntactic issues, data-mapping errors, and hallucinated constraints.

\subsubsection{Automatic Error Decomposition}
\label{subsec:auto_error_decomposition}

Complementing the semantic categories in Figure~\ref{fig:error_analysis}, the
execution harness records automatic failure types such as \wrongvalue,
\runtimeerror, \texttt{module\_not\_found}, and \nameerror.
\textbf{Many failures are executable optimization models with the wrong
objective, not simply broken programs.} Across all $20\times100$ \revise
trials, \wrongvalue accounts for $19.1\%$ of submissions and
\runtimeerror for $18.2\%$. This means that a large share of failures
survive parsing, model construction, and solver invocation, but encode the
wrong mathematical problem. At the same time, \textbf{\revise introduces
legacy-code-specific engineering errors}: \texttt{module\_not\_found} rises
from $0.4\%$ in \build to $2.8\%$ in \revise, and \nameerror rises from
$0.2\%$ to $2.2\%$. These increases are quantitative signatures of context
poisoning, where agents inherit imports, variable names, or control-flow
assumptions from the heuristic without re-grounding them as solver variables
and constraints. Detailed categories are reported in
Appendix~\ref{app:failure_details}.

\subsubsection{Workspace-Specific Failure Modes}
\label{subsec:workspace_failure_modes}

\textbf{The strongest \build model mostly fails on cross-artifact grounding,
not on classical mathematical modelling.} To understand workspace-specific failures beyond automatic labels, we manually classify the 28 \build failures of \texttt{gemini-3.1-pro}, using tracebacks, objective deviations from
$\refobj$, and diffs against \texttt{docs/business\_requirement.md}. Only
$21.4\%$ of failures are mathematical-modelling mistakes, while data-mapping
errors account for $39.3\%$ and hallucinated constraints for $28.6\%$. This
decomposition also shows why the three tasks are not redundant: \build exposes
schema and requirement-grounding errors, \revise exposes legacy-code inheritance
and regression, and \explain exposes fluent but unfaithful reasoning when an
answer fails to cite the actual variable, constraint, unit, slack, or objective
delta recorded in the workspace.

\subsubsection{Case Study: Why Text-Only Evaluation Misses These Errors}
\label{subsec:failure_case_study}

\textbf{A runnable optimal script can still optimize the wrong business
quantity.} In \build instance~2 (\texttt{IndustryOR\_2}), the business document
describes a two-year fighter-pilot training plan. The data fields specify 
\texttt{a1 = 10} fighter jets in year 1, \texttt{a2 = 15} fighter jets in year 2,
\texttt{training\_capacity\_per\_jet = 5} pilots per year, and
\texttt{training\_duration = 2} years, yielding the reference value
$5(10+15)=125$ trained pilots.

\textbf{The error is grounding, not solver execution.}
The \texttt{gemini-3.1-pro} solution was syntactically valid and
returned an \optimalstatus solution, but it introduced combat-jet variables,
maximized \texttt{C1 + C2}, and used
\texttt{C2 <= training\_capacity\_per\_jet * T1}, so only year-1 training jets
could create year-2 pilot capacity. The resulting objective value was 25
instead of 125. The agent mapped \texttt{training\_capacity\_per\_jet} to the
wrong physical quantity and hallucinated a combat-jet objective unsupported by
the workspace. A clean text-only benchmark would typically normalize these
parameters and variables into one prompt, hiding the file, column, unit, and
key-matching step that OR-Space makes explicit.

\section{Conclusion}
\label{sec:conclusion}

In this work, we introduce OR-Space to resolve two critical limitations of existing OR benchmarks: their reliance on textbook-style textual prompts and their focus on single-shot tasks. By unifying the \build, \revise, and \explain phases within a shared multi-artifact workspace, OR-Space evaluates optimization agents across the full industrial OR engineering lifecycle. Our systematic evaluation of 20 language models reveals that navigating an open workspace remains highly challenging, exposing systemic failures in data mapping and constraint grounding that text-only benchmarks largely hide. Ultimately, OR-Space shifts the evaluation paradigm from isolated text generation to robust, solver-grounded engineering workflows, establishing a rigorous foundation for measuring the practical readiness of optimization agents.

\small
\bibliographystyle{plainnat}
\bibliography{refs}

\begin{thebibliography}{47}
\providecommand{\natexlab}[1]{#1}
\providecommand{\url}[1]{\texttt{#1}}
\expandafter\ifx\csname urlstyle\endcsname\relax
  \providecommand{\doi}[1]{doi: #1}\else
  \providecommand{\doi}{doi: \begingroup \urlstyle{rm}\Url}\fi

\bibitem[AhmadiTeshnizi et~al.(2024)AhmadiTeshnizi, Gao, and
  Udell]{ahmaditeshniziOptiMUSScalable2024}
Ali AhmadiTeshnizi, Wenzhi Gao, and Madeleine Udell.
\newblock {OptiMUS}: Scalable optimization modeling with ({MI}){LP} solvers and
  large language models, 2024.
\newblock URL \url{https://arxiv.org/abs/2402.10172}.

\bibitem[Akhtar et~al.(2024)Akhtar, Benjelloun, Conforti, Gijsbers,
  Giner-Miguelez, Jain, Kuchnik, Lhoest, Marcenac, Maskey, Mattson, Oala,
  Ruyssen, Shinde, Simperl, Thomas, Tykhonov, Vanschoren, van~der Velde,
  Vogler, and Wu]{akhtarCroissantMetadataFormat2024}
Mubashara Akhtar, Omar Benjelloun, Costanza Conforti, Pieter Gijsbers, Joan
  Giner-Miguelez, Nitisha Jain, Michael Kuchnik, Quentin Lhoest, Pierre
  Marcenac, Manil Maskey, Peter Mattson, Luis Oala, Pierre Ruyssen, Rajat
  Shinde, Elena Simperl, Goeffry Thomas, Slava Tykhonov, Joaquin Vanschoren,
  Jos van~der Velde, Steffen Vogler, and Carole-Jean Wu.
\newblock Croissant: A metadata format for {ML}-ready datasets.
\newblock In \emph{Proceedings of the Eighth Workshop on Data Management for
  End-to-End Machine Learning (DEEM '24), co-located with SIGMOD/PODS '24},
  pages 1--6, 2024.
\newblock \doi{10.1145/3650203.3663326}.
\newblock URL \url{https://doi.org/10.1145/3650203.3663326}.

\bibitem[Chan et~al.(2024)Chan, Chowdhury, Jaffe, Aung, Sherburn, Mays,
  Starace, Liu, Maksin, Patwardhan, Weng, and M{\k
  a}dry]{chanMLEbenchEvaluatingMachine2024}
Jun~Shern Chan, Neil Chowdhury, Oliver Jaffe, James Aung, Dane Sherburn, Evan
  Mays, Giulio Starace, Kevin Liu, Leon Maksin, Tejal Patwardhan, Lilian Weng,
  and Aleksander M{\k a}dry.
\newblock {MLE-bench}: Evaluating machine learning agents on machine learning
  engineering, 2024.
\newblock URL \url{https://arxiv.org/abs/2410.07095}.
\newblock OpenAI; accepted at ICLR 2025.

\bibitem[Chiang et~al.(2024)Chiang, Zheng, Sheng, Angelopoulos, Li, Li, Zhu,
  Zhang, Jordan, Gonzalez, and Stoica]{chiangChatbotArenaOpen2024}
Wei-Lin Chiang, Lianmin Zheng, Ying Sheng, Anastasios~Nikolas Angelopoulos,
  Tianle Li, Dacheng Li, Banghua Zhu, Hao Zhang, Michael~I. Jordan, Joseph~E.
  Gonzalez, and Ion Stoica.
\newblock Chatbot {Arena}: An open platform for evaluating {LLMs} by human
  preference.
\newblock In \emph{Proceedings of the 41st International Conference on Machine
  Learning (ICML)}, volume 235 of \emph{PMLR}, pages 8359--8388, July 2024.
\newblock URL \url{https://proceedings.mlr.press/v235/chiang24b.html}.

\bibitem[Drouin et~al.(2024)Drouin, Gasse, Caccia, Laradji, Del~Verme, Marty,
  Boisvert, Thakkar, Cappart, Vazquez, Chapados, and
  Lacoste]{drouinWorkArenaCapable2024}
Alexandre Drouin, Maxime Gasse, Massimo Caccia, Issam~H. Laradji, Manuel
  Del~Verme, Tom Marty, L{\'e}o Boisvert, Megh Thakkar, Quentin Cappart, David
  Vazquez, Nicolas Chapados, and Alexandre Lacoste.
\newblock {WorkArena}: How capable are web agents at solving common knowledge
  work tasks?, 2024.
\newblock URL \url{https://arxiv.org/abs/2403.07718}.

\bibitem[Dunning et~al.(2017)Dunning, Huchette, and
  Lubin]{dunningJuMPModelingLanguage2017}
Iain Dunning, Joey Huchette, and Miles Lubin.
\newblock {JuMP}: A modeling language for mathematical optimization.
\newblock \emph{SIAM Review}, 59\penalty0 (2):\penalty0 295--320, 2017.
\newblock \doi{10.1137/15M1020575}.
\newblock URL \url{https://doi.org/10.1137/15M1020575}.

\bibitem[Fourer et~al.(2002)Fourer, Gay, and
  Kernighan]{fourerAMPLModelingLanguage2002}
Robert Fourer, David~M. Gay, and Brian~W. Kernighan.
\newblock \emph{{AMPL}: A Modeling Language for Mathematical Programming}.
\newblock Duxbury Press, 2 edition, 2002.
\newblock URL \url{https://ampl.com/resources/the-ampl-book/}.

\bibitem[Ge et~al.(2022)Ge, Huangfu, Wang, Wu, and
  Ye]{geCardinalOptimizerCOPT2022}
Dongdong Ge, Qi~Huangfu, Zizhuo Wang, Jian Wu, and Yinyu Ye.
\newblock Cardinal optimizer ({COPT}) user guide, 2022.
\newblock URL \url{https://arxiv.org/abs/2208.14314}.

\bibitem[Gebru et~al.(2021)Gebru, Morgenstern, Vecchione, Vaughan, Wallach,
  Daum{\'e}~III, and Crawford]{gebruDatasheetsDatasets2021}
Timnit Gebru, Jamie Morgenstern, Briana Vecchione, Jennifer~Wortman Vaughan,
  Hanna Wallach, Hal Daum{\'e}~III, and Kate Crawford.
\newblock Datasheets for datasets.
\newblock \emph{Communications of the ACM}, 64\penalty0 (12):\penalty0 86--92,
  December 2021.
\newblock \doi{10.1145/3458723}.
\newblock URL \url{https://doi.org/10.1145/3458723}.

\bibitem[{Gurobi Optimization, LLC}(2024)]{gurobi}
{Gurobi Optimization, LLC}.
\newblock Gurobi {Optimizer} {Reference} {Manual}, 2024.
\newblock URL \url{https://www.gurobi.com}.

\bibitem[Hart et~al.(2011)Hart, Watson, and
  Woodruff]{hartPyomoModelingSolving2011}
William~E. Hart, Jean-Paul Watson, and David~L. Woodruff.
\newblock {Pyomo}: Modeling and solving mathematical programs in {Python}.
\newblock \emph{Mathematical Programming Computation}, 3\penalty0 (3):\penalty0
  219--260, 2011.
\newblock \doi{10.1007/s12532-011-0026-8}.
\newblock URL \url{https://doi.org/10.1007/s12532-011-0026-8}.

\bibitem[He et~al.(2026)He, Li, Ji, Liu, and Huang]{heEvoOptLLMEvolving2026}
Yiliu He, Tianle Li, Binghao Ji, Zhiyuan Liu, and Di~Huang.
\newblock {EvoOpt-LLM}: Evolving industrial optimization models with large
  language models, 2026.
\newblock URL \url{https://arxiv.org/abs/2602.01082}.

\bibitem[Huang et~al.(2025{\natexlab{a}})Huang, Tang, Hu, Jiang, Zheng, Ge,
  Wang, and Wang]{huangORLMCustomizableFramework2025}
Chenyu Huang, Zhengyang Tang, Shixi Hu, Ruoqing Jiang, Xin Zheng, Dongdong Ge,
  Benyou Wang, and Zizhuo Wang.
\newblock {ORLM}: {A} {Customizable} {Framework} in {Training} {Large} {Models}
  for {Automated} {Optimization} {Modeling}.
\newblock \emph{Operations Research}, 73\penalty0 (6):\penalty0 2986--3009,
  November 2025{\natexlab{a}}.
\newblock ISSN 0030-364X.
\newblock \doi{10.1287/opre.2024.1233}.
\newblock URL \url{https://pubsonline.informs.org/doi/10.1287/opre.2024.1233}.

\bibitem[Huang et~al.(2026)Huang, Lin, Tang, Jiang, Jiang, Wang, and
  Wei]{huangInvEvolveEvolving2026}
Chenyu Huang, Jianghao Lin, Zhengyang Tang, Bo~Jiang, Ruoqing Jiang, Benyou
  Wang, and Lai Wei.
\newblock {InvEvolve}: Evolving white-box inventory policies via large language
  models with performance guarantees, 2026.
\newblock URL \url{https://arxiv.org/abs/2605.00369}.

\bibitem[Huang et~al.(2025{\natexlab{b}})Huang, Shen, Hu, Gao, and
  Wang]{huangLLMsMathematicalModeling2025}
Xuhan Huang, Qingning Shen, Yan Hu, Anningzhe Gao, and Benyou Wang.
\newblock {LLMs} for mathematical modeling: Towards bridging the gap between
  natural and mathematical languages, 2025{\natexlab{b}}.
\newblock URL \url{https://arxiv.org/abs/2405.13144}.
\newblock Findings of NAACL 2025.

\bibitem[Jimenez et~al.(2024)Jimenez, Yang, Wettig, Yao, Pei, Press, and
  Narasimhan]{jimenezSWEbenchCanLanguage2024}
Carlos~E. Jimenez, John Yang, Alexander Wettig, Shunyu Yao, Kexin Pei, Ofir
  Press, and Karthik Narasimhan.
\newblock {SWE}-bench: {Can} {Language} {Models} {Resolve} {Real}-{World}
  {GitHub} {Issues}?, November 2024.
\newblock URL \url{http://arxiv.org/abs/2310.06770}.
\newblock arXiv:2310.06770 [cs].

\bibitem[Kim et~al.(2023)Kim, Shin, Cho, Jang, Longpre, Lee, Yun, Shin, Kim,
  Thorne, and Seo]{kimPrometheusInducing2023}
Seungone Kim, Jamin Shin, Yejin Cho, Joel Jang, Shayne Longpre, Hwaran Lee,
  Sangdoo Yun, Seongjin Shin, Sungdong Kim, James Thorne, and Minjoon Seo.
\newblock {Prometheus}: Inducing fine-grained evaluation capability in language
  models, 2023.
\newblock URL \url{https://arxiv.org/abs/2310.08491}.
\newblock ICLR 2024.

\bibitem[Li et~al.(2023)Li, Mellou, Zhang, Pathuri, and
  Menache]{liLargeLanguageModels2023}
Beibin Li, Konstantina Mellou, Bo~Zhang, Jeevan Pathuri, and Ishai Menache.
\newblock Large language models for supply chain optimization, 2023.
\newblock URL \url{https://arxiv.org/abs/2307.03875}.

\bibitem[Li et~al.(2026)Li, Lu, Wei, Liu, Chen, Lan, Zhang, and
  Wen]{liConstructingIndustrialScale2026}
Zhong Li, Hongliang Lu, Tao Wei, Wenyu Liu, Yuxuan Chen, Yuan Lan, Fan Zhang,
  and Zaiwen Wen.
\newblock Constructing industrial-scale optimization modeling benchmark, 2026.
\newblock URL \url{https://arxiv.org/abs/2602.10450}.

\bibitem[Liang et~al.(2023)Liang, Bommasani, Lee, Tsipras, Soylu, Yasunaga,
  Zhang, Narayanan, Wu, Kumar, Newman, Yuan, Yan, Zhang, Cosgrove, Manning,
  R{\'e}, Acosta-Navas, Hudson, Zelikman, Durmus, Ladhak, Rong, Ren, Yao, Wang,
  Santhanam, Orr, Zheng, Yuksekgonul, Suzgun, Kim, Guha, Chatterji, Khattab,
  Henderson, Huang, Chi, Xie, Santurkar, Ganguli, Hashimoto, Icard, Zhang,
  Chaudhary, Wang, Li, Mai, Zhang, and
  Koreeda]{liangHolisticEvaluationLanguage2023}
Percy Liang, Rishi Bommasani, Tony Lee, Dimitris Tsipras, Dilara Soylu,
  Michihiro Yasunaga, Yian Zhang, Deepak Narayanan, Yuhuai Wu, Ananya Kumar,
  Benjamin Newman, Binhang Yuan, Bobby Yan, Ce~Zhang, Christian Cosgrove,
  Christopher~D. Manning, Christopher R{\'e}, Diana Acosta-Navas, Drew~A.
  Hudson, Eric Zelikman, Esin Durmus, Faisal Ladhak, Frieda Rong, Hongyu Ren,
  Huaxiu Yao, Jue Wang, Keshav Santhanam, Laurel Orr, Lucia Zheng, Mert
  Yuksekgonul, Mirac Suzgun, Nathan Kim, Neel Guha, Niladri Chatterji, Omar
  Khattab, Peter Henderson, Qian Huang, Ryan Chi, Sang~Michael Xie, Shibani
  Santurkar, Surya Ganguli, Tatsunori Hashimoto, Thomas Icard, Tianyi Zhang,
  Vishrav Chaudhary, William Wang, Xuechen Li, Yifan Mai, Yuhui Zhang, and Yuta
  Koreeda.
\newblock Holistic evaluation of language models.
\newblock \emph{Transactions on Machine Learning Research (TMLR)}, 2023.
\newblock URL \url{https://openreview.net/forum?id=iO4LZibEqW}.
\newblock Featured Certification. arXiv:2211.09110.

\bibitem[Lin et~al.(2026)Lin, Ling, Zhou, Xu, Jiang, Wang, and
  Ge]{linSoliloquyAgoraMemory2026}
Jianghao Lin, Zi~Ling, Chenyu Zhou, Tianyi Xu, Ruoqing Jiang, Zizhuo Wang, and
  Dongdong Ge.
\newblock From soliloquy to agora: Memory-enhanced {LLM} agents with
  decentralized debate for optimization modeling, 2026.
\newblock URL \url{https://arxiv.org/abs/2604.25847}.
\newblock Working paper.

\bibitem[Liu et~al.(2025{\natexlab{a}})Liu, Qin, Huang, Zeng, Xi, Lin, Wu,
  Wang, Shang, Tang, Lian, Yu, and Zhang]{liuRealBarrierLLM2025}
Weiwen Liu, Jiarui Qin, Xu~Huang, Xingshan Zeng, Yunjia Xi, Jianghao Lin,
  Chuhan Wu, Yasheng Wang, Lifeng Shang, Ruiming Tang, Defu Lian, Yong Yu, and
  Weinan Zhang.
\newblock Position: The real barrier to {LLM} agent usability is agentic {ROI},
  2025{\natexlab{a}}.
\newblock URL \url{https://arxiv.org/abs/2505.17767}.

\bibitem[Liu et~al.(2025{\natexlab{b}})Liu, Yu, Zhang, Xu, Lei, Lai, Gu, Ding,
  Men, Yang, Zhang, Deng, Zeng, Du, Zhang, Shen, Zhang, Su, Sun, Huang, Dong,
  and Tang]{liuAgentBenchEvaluatingLLMs2025}
Xiao Liu, Hao Yu, Hanchen Zhang, Yifan Xu, Xuanyu Lei, Hanyu Lai, Yu~Gu,
  Hangliang Ding, Kaiwen Men, Kejuan Yang, Shudan Zhang, Xiang Deng, Aohan
  Zeng, Zhengxiao Du, Chenhui Zhang, Sheng Shen, Tianjun Zhang, Yu~Su, Huan
  Sun, Minlie Huang, Yuxiao Dong, and Jie Tang.
\newblock {AgentBench}: {Evaluating} {LLMs} as {Agents}, October
  2025{\natexlab{b}}.
\newblock URL \url{http://arxiv.org/abs/2308.03688}.
\newblock arXiv:2308.03688 [cs].

\bibitem[Liu et~al.(2023)Liu, Iter, Xu, Wang, Xu, and Zhu]{liuGEvalNLG2023}
Yang Liu, Dan Iter, Yichong Xu, Shuohang Wang, Ruochen Xu, and Chenguang Zhu.
\newblock {G-Eval}: {NLG} evaluation using {GPT-4} with better human alignment.
\newblock In \emph{Proceedings of the 2023 Conference on Empirical Methods in
  Natural Language Processing (EMNLP)}, pages 2511--2522, December 2023.
\newblock \doi{10.18653/v1/2023.emnlp-main.153}.
\newblock URL \url{https://aclanthology.org/2023.emnlp-main.153/}.

\bibitem[Lu et~al.(2025)Lu, Xie, Wu, Ren, Chen, and
  Wen]{luOptMATHScalableBidirectional2025}
Hongliang Lu, Zhonglin Xie, Yaoyu Wu, Can Ren, Yuxuan Chen, and Zaiwen Wen.
\newblock {OptMATH}: A scalable bidirectional data synthesis framework for
  optimization modeling, 2025.
\newblock URL \url{https://arxiv.org/abs/2502.11102}.

\bibitem[Mialon et~al.(2023)Mialon, Fourrier, Swift, Wolf, LeCun, and
  Scialom]{mialonGAIABenchmarkGeneral2023}
Gr{\'e}goire Mialon, Cl{\'e}mentine Fourrier, Craig Swift, Thomas Wolf, Yann
  LeCun, and Thomas Scialom.
\newblock {GAIA}: a benchmark for {General} {AI} {Assistants}, 2023.
\newblock URL \url{https://arxiv.org/abs/2311.12983}.
\newblock arXiv:2311.12983; accepted at ICLR 2024.

\bibitem[Mostajabdaveh et~al.(2025)Mostajabdaveh, Yu, Dash, Ramamonjison,
  Byusa, Carenini, Zhou, and Zhang]{mostajabdavehEvaluatingLLMReasoning2025}
Mahdi Mostajabdaveh, Timothy~T. Yu, Samarendra Chandan~Bindu Dash, Rindranirina
  Ramamonjison, Jabo~Serge Byusa, Giuseppe Carenini, Zirui Zhou, and Yong
  Zhang.
\newblock Evaluating {LLM} reasoning in the operations research domain with
  {ORQA}, 2025.
\newblock URL \url{https://arxiv.org/abs/2412.17874}.
\newblock AAAI 2025.

\bibitem[Pushkarna et~al.(2022)Pushkarna, Zaldivar, and
  Kjartansson]{pushkarnaDataCardsPurposeful2022}
Mahima Pushkarna, Andrew Zaldivar, and Oddur Kjartansson.
\newblock Data {Cards}: Purposeful and transparent dataset documentation for
  responsible {AI}.
\newblock In \emph{Proceedings of the 2022 ACM Conference on Fairness,
  Accountability, and Transparency (FAccT '22)}, pages 1776--1826, 2022.
\newblock \doi{10.1145/3531146.3533231}.
\newblock URL \url{https://doi.org/10.1145/3531146.3533231}.

\bibitem[Ramamonjison et~al.(2023)Ramamonjison, Yu, Li, Li, Carenini, Ghaddar,
  He, Mostajabdaveh, Banitalebi-Dehkordi, Zhou, and
  Zhang]{ramamonjisonNL4OptCompetition2023}
Rindranirina Ramamonjison, Timothy~T. Yu, Raymond Li, Haley Li, Giuseppe
  Carenini, Bissan Ghaddar, Shiqi He, Mahdi Mostajabdaveh, Amin
  Banitalebi-Dehkordi, Zirui Zhou, and Yong Zhang.
\newblock {NL4Opt} competition: Formulating optimization problems based on
  their natural language descriptions, 2023.
\newblock URL \url{https://arxiv.org/abs/2303.08233}.

\bibitem[Stureborg et~al.(2024)Stureborg, Alikaniotis, and
  Suhara]{stureborgLargeLanguageModels2024}
Rickard Stureborg, Dimitris Alikaniotis, and Yoshi Suhara.
\newblock Large language models are inconsistent and biased evaluators, 2024.
\newblock URL \url{https://arxiv.org/abs/2405.01724}.

\bibitem[Thakur et~al.(2024)Thakur, Choudhary, Ramayapally, Vaidyanathan, and
  Hupkes]{thakurJudgingJudges2024}
Aman~Singh Thakur, Kartik Choudhary, Venkat~Srinik Ramayapally, Sankaran
  Vaidyanathan, and Dieuwke Hupkes.
\newblock Judging the judges: Evaluating alignment and vulnerabilities in
  {LLMs}-as-judges, 2024.
\newblock URL \url{https://arxiv.org/abs/2406.12624}.

\bibitem[Trivedi et~al.(2024)Trivedi, Khot, Hartmann, Manku, Dong, Li, Gupta,
  Sabharwal, and Balasubramanian]{trivediAppWorldControllable2024}
Harsh Trivedi, Tushar Khot, Mareike Hartmann, Ruskin Manku, Vinty Dong, Edward
  Li, Shashank Gupta, Ashish Sabharwal, and Niranjan Balasubramanian.
\newblock {AppWorld}: A controllable world of apps and people for benchmarking
  interactive coding agents, 2024.
\newblock URL \url{https://arxiv.org/abs/2407.18901}.
\newblock ACL 2024.

\bibitem[Wang et~al.(2023)Wang, Li, Chen, Cai, Zhu, Lin, Cao, Liu, Liu, and
  Sui]{wangLargeLanguageModels2023}
Peiyi Wang, Lei Li, Liang Chen, Zefan Cai, Dawei Zhu, Binghuai Lin, Yunbo Cao,
  Qi~Liu, Tianyu Liu, and Zhifang Sui.
\newblock Large language models are not fair evaluators, 2023.
\newblock URL \url{https://arxiv.org/abs/2305.17926}.

\bibitem[Xiao et~al.(2024)Xiao, Zhang, Wu, Xu, Wang, Han, Fu, Zhong, Zeng,
  Song, and Chen]{xiaoChainOfExpertsWhen2024}
Ziyang Xiao, Dongxiang Zhang, Yangjun Wu, Lilin Xu, Yuan~Jessica Wang, Xiongwei
  Han, Xiaojin Fu, Tao Zhong, Jia Zeng, Mingli Song, and Gang Chen.
\newblock Chain-of-{Experts}: When {LLMs} meet complex operations research
  problems.
\newblock In \emph{The Twelfth International Conference on Learning
  Representations (ICLR)}, 2024.
\newblock URL \url{https://openreview.net/forum?id=HobyL1B9CZ}.
\newblock Introduces a multi-agent LLM framework for OR modeling and the
  ComplexOR benchmark.

\bibitem[Xie et~al.(2024)Xie, Zhang, Chen, Li, Zhao, Cao, Hua, Cheng, Shin,
  Lei, Liu, Xu, Zhou, Savarese, Xiong, Zhong, and
  Yu]{xieOSWorldBenchmarkingMultimodal2024}
Tianbao Xie, Danyang Zhang, Jixuan Chen, Xiaochuan Li, Siheng Zhao, Ruisheng
  Cao, Toh~Jing Hua, Zhoujun Cheng, Dongchan Shin, Fangyu Lei, Yitao Liu,
  Yiheng Xu, Shuyan Zhou, Silvio Savarese, Caiming Xiong, Victor Zhong, and Tao
  Yu.
\newblock {OSWorld}: Benchmarking multimodal agents for open-ended tasks in
  real computer environments, 2024.
\newblock URL \url{https://arxiv.org/abs/2404.07972}.

\bibitem[Xu et~al.(2024)Xu, Song, Li, Tang, Jain, Bao, Wang, Zhou, Guo, Cao,
  Yang, Lu, Martin, Su, Maben, Mehta, Chi, Jang, Xie, Zhou, and
  Neubig]{xuTheAgentCompanyBenchmarking2024}
Frank~F. Xu, Yufan Song, Boxuan Li, Yuxuan Tang, Kritanjali Jain, Mengxue Bao,
  Zora~Z. Wang, Xuhui Zhou, Zhitong Guo, Murong Cao, Mingyang Yang, Hao~Yang
  Lu, Amaad Martin, Zhe Su, Leander Maben, Raj Mehta, Wayne Chi, Lawrence Jang,
  Yiqing Xie, Shuyan Zhou, and Graham Neubig.
\newblock {TheAgentCompany}: Benchmarking {LLM} agents on consequential real
  world tasks, 2024.
\newblock URL \url{https://arxiv.org/abs/2412.14161}.

\bibitem[Yang et~al.(2025)Yang, Chai, Song, Qi, Wen, Li, Liao, Hu, Lin, Chang,
  Liu, Wen, Yu, and Zhang]{yangSurveyAIAgentProtocols2025}
Yingxuan Yang, Huacan Chai, Yuanyi Song, Siyuan Qi, Muning Wen, Ning Li, Junwei
  Liao, Haoyi Hu, Jianghao Lin, Gaowei Chang, Weiwen Liu, Ying Wen, Yong Yu,
  and Weinan Zhang.
\newblock A survey of {AI} agent protocols, 2025.
\newblock URL \url{https://arxiv.org/abs/2504.16736}.

\bibitem[Yao et~al.(2024)Yao, Shinn, Razavi, and
  Narasimhan]{yaoTauBenchBenchmark2024}
Shunyu Yao, Noah Shinn, Pedram Razavi, and Karthik Narasimhan.
\newblock {$\tau$-bench}: A benchmark for tool-agent-user interaction in
  real-world domains, 2024.
\newblock URL \url{https://arxiv.org/abs/2406.12045}.
\newblock arXiv:2406.12045.

\bibitem[Yin et~al.(2026)Yin, Zhou, Zhu, and Jin]{yinMemDecoderEnhancing2026}
Haoran Yin, Chenyu Zhou, Wei Zhu, and Yuhua Jin.
\newblock {MemDecoder}: Enhancing test-time compute for {LLM} agents via
  reinforced memory decoding.
\newblock In \emph{The Forty-Third International Conference on Machine
  Learning}, 2026.
\newblock URL \url{https://icml.cc/virtual/2026/poster/65523}.

\bibitem[Zhang et~al.(2025{\natexlab{a}})Zhang, Luo, Yang, Soong, and
  Yuen]{zhangORLLMAgentAutomating2025}
Bowen Zhang, Pengcheng Luo, Genke Yang, Boon-Hee Soong, and Chau Yuen.
\newblock {OR-LLM-Agent}: Automating modeling and solving of operations
  research optimization problems with reasoning {LLM}, 2025{\natexlab{a}}.
\newblock URL \url{https://arxiv.org/abs/2503.10009}.

\bibitem[Zhang et~al.(2025{\natexlab{b}})Zhang, Chen, Zope, Barbalho, Mellou,
  Molinaro, Kulkarni, Menache, and Li]{zhangOptiMindTeaching2025}
Xinzhi Zhang, Zeyi Chen, Humishka Zope, Hugo Barbalho, Konstantina Mellou,
  Marco Molinaro, Janardhan Kulkarni, Ishai Menache, and Sirui Li.
\newblock {OptiMind}: Teaching {LLMs} to think like optimization experts,
  2025{\natexlab{b}}.
\newblock URL \url{https://arxiv.org/abs/2509.22979}.

\bibitem[Zheng et~al.(2023)Zheng, Chiang, Sheng, Zhuang, Wu, Zhuang, Lin, Li,
  Li, Xing, Zhang, Gonzalez, and Stoica]{zheng2024judging}
Lianmin Zheng, Wei-Lin Chiang, Ying Sheng, Siyuan Zhuang, Zhanghao Wu, Yonghao
  Zhuang, Zi~Lin, Zhuohan Li, Dacheng Li, Eric~P. Xing, Hao Zhang, Joseph~E.
  Gonzalez, and Ion Stoica.
\newblock Judging {LLM}-as-a-{Judge} with {MT}-{Bench} and {Chatbot} {Arena}.
\newblock In \emph{Advances in Neural Information Processing Systems (NeurIPS),
  Datasets and Benchmarks Track}, 2023.
\newblock URL \url{https://arxiv.org/abs/2306.05685}.

\bibitem[Zhou et~al.(2025{\natexlab{a}})Zhou, Xu, Lin, and
  Ge]{zhouStepORLMSelfEvolving2025}
Chenyu Zhou, Tianyi Xu, Jianghao Lin, and Dongdong Ge.
\newblock {StepORLM}: A self-evolving framework with generative process
  supervision for operations research language models, 2025{\natexlab{a}}.
\newblock URL \url{https://arxiv.org/abs/2509.22558}.

\bibitem[Zhou et~al.(2025{\natexlab{b}})Zhou, Yang, Xin, Chen, He, and
  Ge]{zhouAutoFormulatingDynamic2025}
Chenyu Zhou, Jingyuan Yang, Linwei Xin, Yitian Chen, Ziyan He, and Dongdong Ge.
\newblock Auto-formulating dynamic programming problems with large language
  models, 2025{\natexlab{b}}.
\newblock URL \url{https://arxiv.org/abs/2507.11737}.

\bibitem[Zhou et~al.(2026)Zhou, Chai, Chen, Guo, Shan, Song, Xu, Yang, Yu,
  Zhang, Zheng, Zhu, Zheng, Zhang, Lou, Zhang, Fu, Wang, Liu, Lin, and
  Zhang]{zhouExternalizationLLMAgents2026}
Chenyu Zhou, Huacan Chai, Wenteng Chen, Zihan Guo, Rong Shan, Yuanyi Song,
  Tianyi Xu, Yingxuan Yang, Aofan Yu, Weiming Zhang, Congming Zheng, Jiachen
  Zhu, Zeyu Zheng, Zhuosheng Zhang, Xingyu Lou, Changwang Zhang, Zhihui Fu, Jun
  Wang, Weiwen Liu, Jianghao Lin, and Weinan Zhang.
\newblock Externalization in {LLM} agents: A unified review of memory, skills,
  protocols and harness engineering, 2026.
\newblock URL \url{https://arxiv.org/abs/2604.08224}.

\bibitem[Zhou et~al.(2024)Zhou, Xu, Zhu, Zhou, Lo, Sridhar, Cheng, Ou, Bisk,
  Fried, Alon, and Neubig]{zhouWebArenaRealisticWeb2024}
Shuyan Zhou, Frank~F. Xu, Hao Zhu, Xuhui Zhou, Robert Lo, Abishek Sridhar,
  Xianyi Cheng, Tianyue Ou, Yonatan Bisk, Daniel Fried, Uri Alon, and Graham
  Neubig.
\newblock {WebArena}: {A} {Realistic} {Web} {Environment} for {Building}
  {Autonomous} {Agents}, April 2024.
\newblock URL \url{http://arxiv.org/abs/2307.13854}.
\newblock arXiv:2307.13854 [cs].

\bibitem[Zhu et~al.(2025)Zhu, Zhu, Rui, Shan, Zheng, Chen, Xi, Lin, Liu, Tang,
  Yu, and Zhang]{zhuEvolutionaryPerspectivesEvaluation2025}
Jiachen Zhu, Menghui Zhu, Renting Rui, Rong Shan, Congmin Zheng, Bo~Chen,
  Yunjia Xi, Jianghao Lin, Weiwen Liu, Ruiming Tang, Yong Yu, and Weinan Zhang.
\newblock Evolutionary perspectives on the evaluation of {LLM}-based {AI}
  agents: A comprehensive survey, 2025.
\newblock URL \url{https://arxiv.org/abs/2506.11102}.

\end{thebibliography}

\appendix

\newpage

\section*{Appendix}
{
\linespread{1.5}\selectfont
\setlength{\parskip}{0pt}
\def\apptocline#1#2#3{%
  \noindent\hyperref[#3]{#1\hspace{0.6em}#2}%
  \leaders\hbox to 0.55em{\hss.\hss}\hfill\hyperref[#3]{\pageref*{#3}}\par}
\def\apptocsubline#1#2#3{%
  \noindent\hspace{1.5em}\hyperref[#3]{#1\hspace{0.6em}#2}%
  \leaders\hbox to 0.55em{\hss.\hss}\hfill\hyperref[#3]{\pageref*{#3}}\par}
\apptocline{A}{Additional Related Work}{app:additional_related}
\apptocline{B}{Benchmark Construction and Reproducibility}{app:benchmark_construction}
\apptocsubline{B.1}{Dataset Construction and Prompting Templates}{appendix:prompts}
\apptocsubline{B.2}{Release, Reproducibility, and Responsible Use}{app:release_responsible_use}
\apptocline{C}{Additional Experimental Results}{app:additional_experimental_results}
\apptocsubline{C.1}{Full Solver-Backend Results}{app:full_solver_backend_results}
\apptocsubline{C.2}{Filesystem Interface Delta Details}{appendix:fs_vs_flat}
\apptocsubline{C.3}{Solver-Backend Sensitivity Details}{appendix:solver_comparison}
\apptocsubline{C.4}{\revise Context Model-Level Details}{appendix:revise_context_ablation}
\apptocsubline{C.5}{Failure Details and Case Studies}{app:failure_details}
\apptocline{D}{Supplementary Dataset Statistics and Visualizations}{app:supp_plots}
\apptocline{E}{\explain Evaluation Rubric}{appendix:rubric}
}
\vspace{0.2em}

\section{Additional Related Work}
\label{app:additional_related}

\paragraph{Optimization modeling benchmarks.}
NL4Opt frames optimization modeling as semantic parsing from natural language to solver-ready linear programming representations~\cite{ramamonjisonNL4OptCompetition2023}. Mamo uses solvers to evaluate whether LLM-generated mathematical models produce correct numerical results across ordinary differential equations, linear programming, and mixed-integer linear programming~\cite{huangLLMsMathematicalModeling2025}. ORLM studies data synthesis and open-source model training for optimization modeling through OR-Instruct and IndustryOR~\cite{huangORLMCustomizableFramework2025}. OptiMUS decomposes modeling into formulation, solver-code generation, debugging, solution evaluation, and refinement~\cite{ahmaditeshniziOptiMUSScalable2024}, while Chain-of-Experts coordinates multiple LLM experts and releases ComplexOR~\cite{xiaoChainOfExpertsWhen2024}. Recent datasets and training frameworks extend this direction: OptMATH scales bidirectional data synthesis for optimization modeling~\cite{luOptMATHScalableBidirectional2025}; DP-Bench/DPLM studies auto-formulation for dynamic programming~\cite{zhouAutoFormulatingDynamic2025}; ORQA evaluates domain-specific OR reasoning and OptiGuide studies supply-chain optimization explanation~\cite{mostajabdavehEvaluatingLLMReasoning2025,liLargeLanguageModels2023}; OptiMind and StepORLM inject optimization expertise, solver verification, and process supervision into model training~\cite{zhangOptiMindTeaching2025,zhouStepORLMSelfEvolving2025}; OR-LLM-Agent, EvoOpt-LLM, InvEvolve, and Agora-Opt study agentic decomposition, model or policy evolution, memory, and debate for OR modeling and deployment~\cite{zhangORLLMAgentAutomating2025,heEvoOptLLMEvolving2026,huangInvEvolveEvolving2026,linSoliloquyAgoraMemory2026}; and MIPLIB-NL stresses industrial scale and model--data separation from real MIP instances~\cite{liConstructingIndustrialScale2026}. OR-Space is complementary: it retains solver-grounded evaluation, but moves from isolated formulations to persistent workspaces with documents, data, code, and execution state.

\paragraph{Interactive agent benchmarks.}
SWE-bench evaluates repository editing against tests~\cite{jimenezSWEbenchCanLanguage2024}; WebArena evaluates browser agents in realistic websites~\cite{zhouWebArenaRealisticWeb2024}; AgentBench spans multiple interactive environments~\cite{liuAgentBenchEvaluatingLLMs2025}; GAIA uses real files and tools for general-assistant tasks~\cite{mialonGAIABenchmarkGeneral2023}; MLE-bench evaluates Kaggle-style machine-learning engineering~\cite{chanMLEbenchEvaluatingMachine2024}; and $\tau$-bench studies multi-turn tool-agent-user interaction under domain policies, with broader survey and ROI-centered perspectives emphasizing environment, metrics, and usability costs~\cite{yaoTauBenchBenchmark2024,zhuEvolutionaryPerspectivesEvaluation2025,liuRealBarrierLLM2025}. OSWorld evaluates multimodal agents in real computer environments with file I/O and execution-based checks~\cite{xieOSWorldBenchmarkingMultimodal2024}; WorkArena targets enterprise knowledge-work tasks in web software~\cite{drouinWorkArenaCapable2024}; AppWorld evaluates interactive coding agents over realistic app APIs with state-based unit tests~\cite{trivediAppWorldControllable2024}; and TheAgentCompany simulates a workplace where agents browse, code, run programs, and communicate~\cite{xuTheAgentCompanyBenchmarking2024}. Work on agent externalization, protocols, and memory selection provides a systems lens on why memory, skills, protocols, and harness design can change what an LLM agent is able to do in practice~\cite{zhouExternalizationLLMAgents2026,yangSurveyAIAgentProtocols2025,yinMemDecoderEnhancing2026}. OR-Space differs in its oracle and state: correctness is mathematical consistency among requirements, parameters, variables, constraints, objectives, and returned solutions.

\paragraph{Benchmark validity and judge design.}
HELM argues for multi-metric reporting rather than a single proxy for general capability~\cite{liangHolisticEvaluationLanguage2023}. Dataset documentation work such as Datasheets for Datasets~\cite{gebruDatasheetsDatasets2021}, Data Cards~\cite{pushkarnaDataCardsPurposeful2022}, and Croissant~\cite{akhtarCroissantMetadataFormat2024} motivates explicit reporting of motivation, composition, collection, and intended use. OR-Space follows this practice by reporting \build, \revise, and \explain separately and tying each to its oracle. For \explain, we use an LLM-as-judge only for short workspace-grounded responses; the rubric is explicit to mitigate sensitivity to judge prompts, positional effects, anchoring, leniency, and other known evaluator biases~\cite{zheng2024judging,chiangChatbotArenaOpen2024,liuGEvalNLG2023,kimPrometheusInducing2023,wangLargeLanguageModels2023,stureborgLargeLanguageModels2024,thakurJudgingJudges2024}.

\section{Benchmark Construction and Reproducibility}
\label{app:benchmark_construction}

\subsection{Dataset Construction and Prompting Templates}
\label{appendix:prompts}

This section records, verbatim, the prompt templates used by every LLM call in
OR-Space. They cover three groups: (i)~\textbf{dataset forging} prompts (\build
lift, \revise-M L5 forge, Business-voice rewrite), (ii)~\textbf{quality gates}
(difficulty judge, business-voice 5-dim rubric judge, open-ended rubric judge),
and (iii)~\textbf{evaluation} prompts (agents solving \build/\revise workspaces,
\explain item authors). All templates are released alongside the code; the
placeholder \texttt{\{...\}} fields are filled in at runtime from the
workspace instance.

All calls use deterministic decoding (temperature $\le 0.3$; $0.0$ for
judges) and \texttt{response\_format=json\_object} whenever a strict JSON
return schema is declared.

Before listing the prompts, we summarize the on-disk layout that these
templates operate on. The benchmark is organized around three workspace
families, each containing 100 instances: \texttt{build\_workspaces/},
\texttt{revise\_workspaces/}, and \texttt{explain\_workspaces/}. The \build
setting exposes a single workspace with data, docs, and source code;
\revise splits the problem into \texttt{original/} and \texttt{revised/}
subdirectories; \explain follows the same two-stage structure but additionally
records solver logs and solver artifacts for both versions.

\begin{figure}[!htbp]
\centering
\includegraphics[width=1\textwidth]{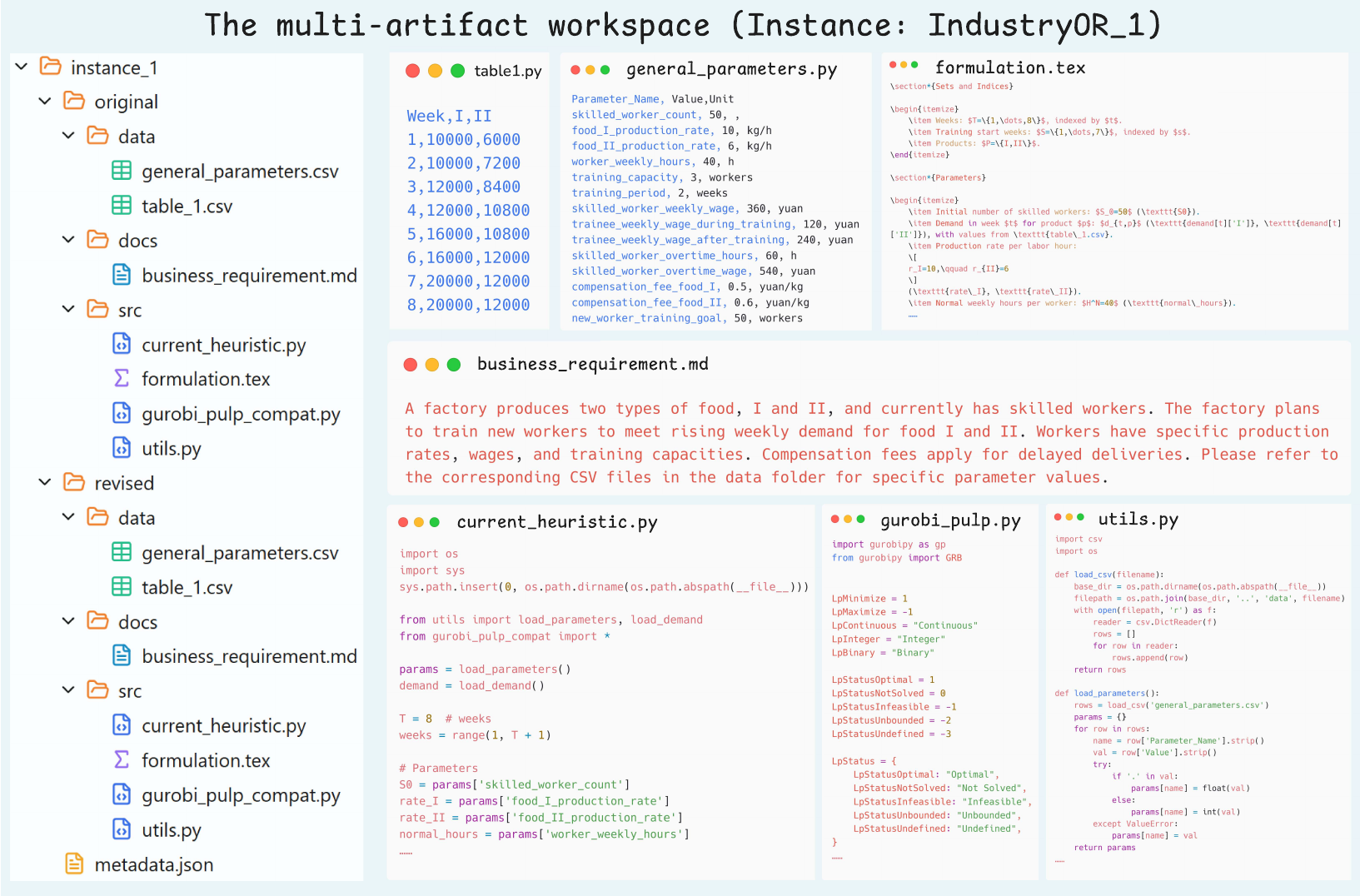}
\caption{Workspace anatomy of \texttt{IndustryOR\_1}. The left panel shows the workspace file hierarchy, while the right panels illustrate the distributed context an agent must synthesize across documents, data tables, and legacy code.}
\label{fig:workspace_anatomy}
\end{figure}

To concretely illustrate the workspace-based architecture of OR-Space, Figure~\ref{fig:workspace_anatomy} presents a high-fidelity visualization of the \texttt{IndustryOR\_1} instance. This example highlights the distributed cross-artifact context that agents must navigate when performing industrial OR tasks.

\begin{tcolorbox}[
  title={Representative workspace layout},
  colback=gray!4,
  colframe=black!35,
  boxrule=0.4pt,
  left=4pt,
  right=4pt,
  top=4pt,
  bottom=4pt,
  breakable
]
\footnotesize
\begin{verbatim}
Workspace_OR/
|-- build_workspaces/
|   |-- instance_k/
|   |   |-- metadata.json
|   |   |-- _agent_solution.py
|   |   |-- data/    (general_parameters.csv, table_1.csv, ...)
|   |   |-- docs/    (business_requirement.md)
|   |   `-- src/     (current_heuristic.py, formulation.tex,
|   |                 gurobi_pulp_compat.py, utils.py)
|   `-- instance_1 ... instance_100
|
|-- revise_workspaces/
|   |-- instance_k/
|   |   |-- metadata.json
|   |   |-- original/ (data/, docs/, src/)
|   |   `-- revised/  (data/, docs/, src/)
|   `-- instance_1 ... instance_100
|
`-- explain_workspaces/
    |-- index.json
    |-- instance_k/
    |   |-- metadata.json
    |   |-- original/
    |   |   |-- data/  docs/  src/
    |   |   |-- logs/           (execution_record.json, solver_output.log)
    |   |   `-- solver_records/ (Gurobi log, LP relaxation, solution JSON)
    |   `-- revised/
    |       |-- data/  docs/  src/
    |       |-- logs/
    |       `-- solver_records/
    `-- instance_1 ... instance_100
\end{verbatim}
\end{tcolorbox}

This representative tree is intentionally shown at the instance level rather
than enumerating all 300 task directories. Instance-specific names can vary
slightly across domains, but the artifact roles are fixed: \texttt{docs/}
contains the natural-language requirement, \texttt{data/} stores all numeric
parameters, \texttt{src/} contains modelling code or reference heuristics,
\texttt{logs/} captures execution traces, and \texttt{solver\_records/}
stores solver-side evidence used by \explain questions.

The remainder of this appendix follows the same operational order as the
dataset pipeline. We begin with the prompts used to forge the workspaces,
continue with the quality gates that filter and rewrite candidate instances,
and finish with the evaluation prompts used at benchmark time.

\subsubsection{Dataset Forging Prompts}

\paragraph{P1. Text $\rightarrow$ Workspace (\build lift).}
Given a raw IndustryOR text instance we ask an author LLM to emit a complete
workspace~$(\docset,\paramset,\codeset)$ as strict JSON. The key
constraint is that $\codeset$ is split into a two-file solver
(\texttt{current\_heuristic.py} + \texttt{utils.py}) that reads CSV data from
\texttt{../data/} and prints a single line \texttt{OBJECTIVE\_VALUE:~<v>}.
\begin{promptbox}{P1 System prompt (\build lift)}
You are an OR workspace author. Convert a clean OR problem statement into an
executable workspace with three folders: docs/, data/, src/.\\
\\
Strict requirements:\\
1. docs/business\_requirement.md restates the problem in business voice; every numeric parameter is quoted but NOT duplicated as a raw table.\\
2. data/*.csv hold all numeric parameters. Use general\_parameters.csv for scalars and table\_\{k\}.csv for indexed data.\\
3. src/current\_heuristic.py reads CSVs from ../data/, builds a PuLP model, solves with commercial solver backends (e.g., via pulp.GUROBI\_CMD or pulp.COPT), and prints the single final line OBJECTIVE\_VALUE: <value>.\\
4. Helper math (e.g. big-M derivation, index construction) lives in src/utils.py and is imported by current\_heuristic.py.\\
5. Running cd src \&\& python current\_heuristic.py must reproduce the ground-truth objective within $10^{-3}$ relative tolerance.\\
\\
Return strict JSON:\\
\{\\
\quad "docs": \{...\},\\
\quad "data": \{...\},\\
\quad "src": \{...\},\\
\quad "run": \{\\
\quad \quad "run.sh": "cd src \&\& python current\_heuristic.py"\\
\quad \},\\
\quad "evaluation": \{\\
\quad \quad "ground\_truth": <float>,\\
\quad \quad "tolerance": 0.01\\
\quad \}\\
\}
\end{promptbox}

The $10^{-3}$ condition in P1 is a generation-time quality gate used to accept
workspace drafts; the released scoring metadata records the benchmark
tolerance field used by the objective oracle.

\paragraph{P2. \revise-Modeling L5 forge.}
The L5 \revise forge is a two-call protocol (author + independent verifier).
The author prompt embeds the full L5 specification: a quantitative operator
profile (counts of $+\!var$, $-\!var$, $+c$, $-c$, $\text{mod\,}c$, $\text{mod\,}\!obj$, $\text{mod\,data}$), an ``L5 iff'' criterion, and three worked templates
(multi-stage decisions with shared resources, linked if-then constraints,
piecewise-linear objective transform). A condensed form of the author prompt
is shown below; the full text is released as
\texttt{prompts/regenerate\_revise\_prompt.md}.

\begin{promptbox}{P2 System prompt (\revise-M L5 forge, condensed)}
You generate \textbf{one} L5 revise of a given OR instance.\\
\\
Design principles (all hard):\\
$\bullet$ Business need must be \emph{unambiguous}: every new parameter has a numeric value in a CSV and a verbal definition in business\_requirement.md; no ``reasonable values'', no unnamed penalties.\\
$\bullet$ Difficulty arises from structure, not prose opacity. Trigger at least one of \{coupling, cascading, implicit\_dependency\}.\\
$\bullet$ L5 iff $\text{distinct\_ops} \ge 4$ OR $(\text{coupling} \wedge \text{cascading} \wedge \text{op\_count} \ge 5)$.\\
$\bullet$ Revised GT must differ from the original; the problem stays feasible and bounded; CBC solves within 60\,s.\\
$\bullet$ Start revise\_description with a \emph{business motivation} sentence; write $\le 250$ chars; reference every new parameter by name.\\
\\
Return a strict JSON with the following structure:\\
\{\\
\quad "instance\_id": ...,\\
\quad "revise\_type\_name": ...,\\
\quad "revise\_description": ...,\\
\quad "metadata": \{\\
\quad \quad "diff": \{ "operator\_counts": ..., "booleans": ..., "complexity\_level": ... \}\\
\quad \},\\
\quad "original\_workspace": \{ "docs": ..., "data": ..., "src": ... \},\\
\quad "revised\_workspace": \{ "docs": ..., "data": ..., "src": ... \},\\
\quad "evaluation": \{\\
\quad \quad "original\_ground\_truth": ...,\\
\quad \quad "revised\_ground\_truth": ...\\
\quad \}\\
\}\\
\\
Follow templates A (multi-stage + shared resource), B (activation + if-then), or C (piecewise objective + variable collapse); do not output commentary.
\end{promptbox}

\begin{promptbox}{P2' Verifier prompt (independent re-solve)}
You are an independent OR modeller.Solve \emph{only} from
\texttt{docs/business\_requirement.md} and \texttt{data/*.csv}. You may NOT
read the author's \texttt{src/}. Produce your own PuLP script, print
\texttt{OBJECTIVE\_VALUE: <v>}, and stop. Your objective must agree with the
author's revised ground truth within $10^{-3}$ relative tolerance; otherwise
the instance is rejected.
\end{promptbox}

As in P1, the $10^{-3}$ verifier threshold is used during dataset construction
to reject inconsistent generated workspaces; benchmark submissions are scored
with the objective tolerance stated in Section~\ref{sec:experiments}.

\paragraph{P3. Business-voice rewrite.}
The \emph{mechanical} revised description (which explicitly names
constraints, penalties, and variables) is rewritten into an authentic
business-voice version, producing the $W_{\text{rev-b}}$ workspace. The
rewrite must preserve every numeric parameter and CSV name while removing all
meta-language.

\begin{promptbox}{P3 System prompt (Business-voice rewrite)}
You are a business analyst documenting an operational change for an OR team.
Rewrite the \emph{revised} \texttt{business\_requirement.md} so it reads like
its build-time sibling (same voice, same paragraph style).\\[0.3em]
\textbf{Forbidden meta-language}: ``add a constraint'', ``modify the
objective'', ``introduce a variable'', ``penalize \ldots'', ``this revision'',
``the model must'', ``subject to'', ``in addition to the original problem'',
explicit MILP / big-M / decision-variable / objective-function vocabulary.\\
\textbf{Forbidden CSV leaks}: backticked parameter names
(\texttt{regular\_hours\_limit\_day\_A}, \texttt{bigM\_shift\_A},
\texttt{z\_target}, \texttt{penalty\_*}, etc.). CSV file names such as
\texttt{general\_parameters.csv} are allowed because the build voice already
references them.\\
\textbf{Forbidden corporate clich\'e}: ``network operations team'', ``recent
post-incident review'', ``ESG'', ``compliance review'', ``management wants''.\\
\textbf{Forbidden model-translator shorthand}: ``fairness penalty'', ``fixed
charge'', ``per-unit penalty'', ``a setup cost is incurred''.\\[0.3em]
\textbf{Encouraged}: a credible operational story that explains \emph{why}
the rule exists; let the modeller turn the story into a constraint or cost.
All numeric values must be retained verbatim.\\
Length budget: $1.0\times$ to $3.0\times$ of the \emph{original build}
description (not of the mechanical revised one).
\end{promptbox}

\subsubsection{Quality-Gate Prompts}

\paragraph{P4. Three-axis difficulty judge.}
Each $(W_{\text{orig}}, W_{\text{rev}})$ pair is scored by an LLM judge on
six archetypes (A--F) with score caps and three 0--5 axes: coupling,
cascading, implicit dependency.

\begin{promptbox}{P4 System prompt (Difficulty judge)}
You are a rubric difficulty judge for OR revise tasks. Read the original
workspace, the revised workspace, and the computed diff. Return STRICT JSON:\\
\texttt{\{"archetype":"A|B|C|D|E|F",}\\
\texttt{~"coupling":0..5, "cascading":0..5, "implicit\_dependency":0..5,}\\
\texttt{~"composite":<sum>, "tier":"L1|L2|L3|L4|L5", "justification":"..."\}}\\[0.3em]
Archetype caps (hard):\\
A (Parameter tweak) $\le 2$ on each axis; B (Local bound) $\le 3$;
C (Business rule) $\le 4$; D (Penalty/fixed cost) $\le 4$;
E (Objective transform) $\le 5$; F (Scenario restructuring) $\le 5$.\\
Tier: composite $\le 4 \Rightarrow$ L1; $5$--$7$ L2; $8$--$10$ L3;
$11$--$13$ L4; $\ge 14$ L5.\\
Axes definitions: \emph{coupling} = a new/removed variable forces existing
constraints to change; \emph{cascading} = one edit propagates through the
existing model topology; \emph{implicit\_dependency} = the change is stated
via business semantics and must be inferred into a structural edit.
\end{promptbox}

\paragraph{P5. Business-voice 5-dim rubric judge.}
Each rewritten instance is scored $1$--$5$ on five axes
(D1~Style, D2~No-Meta-Guidance, D3~Numerical Specificity, D4~Motivation,
D5~Inference Difficulty). Instances below $4$ on D1--D4 go through the
rescue loop (P3 re-run).
\begin{promptbox}{P5 System prompt (5-dim business-voice rubric)}
You evaluate a REVISED business description against the ORIGINAL build-style description of the same problem. For each of the five dimensions below, return an integer score 1--5 and a one-sentence justification.\\
\\
Evaluation Dimensions:\\
\quad \textbf{D1 Style}: does it sound like the build-time sibling document (tone, paragraph shape, level of formality)?\\
\quad \textbf{D2 No-Meta-Guidance}: does it avoid instructing the modeller (no ``add a constraint'', no ``penalize ...'', no parameter backticks, no MILP vocabulary)?\\
\quad \textbf{D3 Numerical Specificity}: are every numeric limit, ratio, and threshold from the mechanical revise retained and used correctly?\\
\quad \textbf{D4 Motivation}: is there a credible operational reason for the change (supplier renegotiation, rota policy, etc.)?\\
\quad \textbf{D5 Inference Difficulty}: after reading, does the modeller still have to \emph{infer} the constraint/cost? (Score low if the text paraphrases the equation; score high if authentic inference is required.)\\
\\
Return STRICT JSON:\\
\{\\
\quad "D1": n,\\
\quad "D2": n,\\
\quad "D3": n,\\
\quad "D4": n,\\
\quad "D5": n,\\
\quad "notes": "..."\\
\}
\end{promptbox}

\paragraph{P6. Open-ended rubric judge (\explain).}
Each open-ended \explain item has three atomic anchors (numeric, entity,
causal). A regex pass tries a cheap automatic hit; numeric regex hits still
go through a semantic verification by the LLM judge.

\begin{promptbox}{P6 System prompt (\explain open-ended judge)}
You are a strict RUBRIC JUDGE for operations-research open-ended answers.
You will receive (i) the question, (ii) a candidate ANSWER, (iii) a single
RUBRIC ANCHOR (one atomic fact).\\[0.3em]
Return STRICT JSON \texttt{\{"hit":0|1,"reason":"<=30 words"\}} under the
following rules:\\
1. HIT=1 only if the anchor fact is present AND used in the right context;
a coincidental number inside an unrelated phrase is NOT a hit.\\
2. MISS=0 if the fact is absent, negated, or only vaguely gestured at.\\
3. Strict on numeric anchors: the exact number (or an algebraically
equivalent expression) must appear.\\
4. Lenient on surface form: synonyms / paraphrases / symbolic notation are
accepted if meaning is identical.
\end{promptbox}

\subsubsection{Evaluation Prompts}

\paragraph{P7. \build/\revise-M code evaluator.}
For the headline benchmark we ask the model-under-test to produce a
solver-agnostic PuLP script that exposes \texttt{build\_problem()
$\to$ pulp.LpProblem} (no solve-call). The runner then attaches
the configured solver backend and records the solved objective value.

\begin{promptbox}{P7 System prompt (\build / \revise-M solver code)}
You are an OR-modelling Python expert.\\
Given (A) the original problem, (B) the revise description, (C) the revised
problem and data, produce a COMPLETE, SELF-CONTAINED Python script that
models the REVISED problem using \texttt{pulp}.\\
1. Read data ONLY from \texttt{./data/<filename>}.\\
2. Use \texttt{pulp}, \texttt{pandas}, and the Python standard library only.\\
3. Define \texttt{build\_problem()} that returns a populated
\texttt{pulp.LpProblem}. Do NOT call \texttt{prob.solve()} inside it.\\
4. At module top level, define \texttt{PROBLEM = build\_problem()}.\\
5. The runner attaches different solvers; just produce the model.\\
6. Return ONLY Python source code. No markdown fences, no commentary, no
language tag, no JSON.
\end{promptbox}

\paragraph{P8. \revise-B workspace agent.}
The \revise-B setting materialises the workspace on disk; the agent sees
\texttt{./docs/}, \texttt{./data/}, \texttt{./src/} and must write a
\emph{new} \texttt{user\_model.py}. Compared to P7, the input is the
business-voice description (no meta-language) plus the original
\texttt{current\_heuristic.py} for reference.

\begin{promptbox}{P8 System prompt (\revise-B on-disk workspace)}
You are an OR-modelling Python expert operating inside an OR-Space workspace.\\
Layout:\\
\hspace*{1em}\texttt{./docs/business\_requirement.md}~~(REVISED problem)\\
\hspace*{1em}\texttt{./docs/REVISE\_NOTE.md}~~(short change note)\\
\hspace*{1em}\texttt{./data/<csv files>}~~(revised data, already in place)\\
\hspace*{1em}\texttt{./src/current\_heuristic.py}~~(original code, reference only)\\[0.3em]
Write a NEW self-contained script \texttt{user\_model.py} that models the
REVISED problem.\\
Strict:\\
(1) Read data ONLY from \texttt{./data/<filename>}.\\
(2) Use \texttt{pulp}, \texttt{pandas}, \texttt{numpy}, \texttt{csv}, and
stdlib only.\\
(3) Expose \texttt{build\_problem() $\to$ pulp.LpProblem} AND
module-level \texttt{PROBLEM}.\\
(4) Do NOT call \texttt{solve()}; the harness attaches the configured solver backend.\\
(5) If data files have renamed/dropped columns, adapt the loader; do not
blindly copy the original column names.\\
(6) Return ONLY raw Python source. No fences, no prose.
\end{promptbox}

\paragraph{P9. \explain item authors.}
\explain MCQs are written by three parallel authors, one per dimension. Each
author enforces the same 4-distractor personality set
(Correct / Plausible-but-Incomplete / Academic-Buzzword-Trap /
Logical-Inverse) and must emit a \texttt{reasoning\_explanation} that only
the \emph{correct} option survives a multi-hop trace
$\docset\to\codeset\to\paramset\to\metricset$.

\begin{promptbox}{P9a Dim-1 author (cross-file entity alignment)}
Design a 4-option question asking what a CODE expression physically represents. The correct answer must RECONCILE business\_requirement.md (business meaning) + code (mathematical definition) + CSV (units).\\
\\
Distractor personalities:\\
\quad A. \textbf{Correct}: exact physical quantity with correct unit, reconciling md + py + csv.\\
\quad B. \textbf{Plausible but incomplete}: looks only at code OR only at doc; misses the CSV unit subtlety.\\
\quad C. \textbf{Academic buzzword trap}: invokes an OR term (network flow, KKT stationarity, dual simplex) that does NOT apply.\\
\quad D. \textbf{Logical inverse}: swaps two entities (I vs II, capacity vs demand) or flips direction.\\
\\
Return strict JSON:\\
\{\\
\quad "question": "...",\\
\quad "options": \{\\
\quad \quad "A": "...",\\
\quad \quad "B": "...",\\
\quad \quad "C": "...",\\
\quad \quad "D": "..."\\
\quad \},\\
\quad "reasoning\_explanation": "..."\\
\}\\
\end{promptbox}

\begin{promptbox}{P9b Dim-2 author (dual / binding / slack)}
Based on the SOLVER FACTS (constraint name, RHS, LP dual, slack, binding flag,
MIP vs LP objectives, integer gap, problem sense), design a 4-option question
on shadow-price meaning and its implication for a $\pm 1$ change in RHS.\\
Distractor rules:\\
A Correct: integrates slack=0 + dual sign + LP$\to$MIP caveat correctly.\\
B Plausible but Incomplete: correct for LP but IGNORES the integer gap or the
shadow-price validity range.\\
C Academic Buzzword Trap: misuses KKT / complementary slackness / Benders
terminology.\\
D Logical Inverse: flips the sign of the dual OR the direction of RHS change.
\end{promptbox}

\begin{promptbox}{P9c Dim-3 author (\revise linkage + big-M tightness)}
Given (orig workspace, revise diff, revised workspace, revised MIP state),
design a 4-option question asking what structural effect the revision has on
the feasible region or on a chosen big-M.\\
Correct answer must trace a causal chain
\texttt{revise\_description} $\to$ added/removed variable $\to$ existing
constraint that tightens $\to$ binding/slack change. Distractors B/C/D use
the same personality roles as P9a.
\end{promptbox}

\paragraph{P10. Open-ended lifter.}
Each MCQ is lifted into a free-form question with exactly three rubric
anchors. The lifter prompt below is deterministic and declares the anchor
schema explicitly, to keep P6 judging reliable.

\begin{promptbox}{P10 System prompt (MCQ $\to$ open-ended lifter)}
Convert the MCQ (question, correct option A) into a free-form question that elicits the same reasoning without any options.\\
\\
Also emit exactly THREE rubric anchors that a correct answer must contain:\\
\quad 1. \textbf{numeric}: an exact value (with unit) that must appear;\\
\quad 2. \textbf{entity}: the correct variable, constraint name, or business term;\\
\quad 3. \textbf{causal}: the reasoning link (``because ... therefore ...'').\\
\\
Return strict JSON:\\
\{\\
\quad "question": "...",\\
\quad "gold\_answer": "...",\\
\quad "rubric\_anchors": [\\
\quad \quad \{ "type": "numeric", "text": "...", "regex": "..." \},\\
\quad \quad \{ "type": "entity", "text": "...", "regex": "..." \},\\
\quad \quad \{ "type": "causal", "text": "...", "regex": null \}\\
\quad ]\\
\}\\
\\
The regex is optional and used for a cheap automatic hit; the final decision is always made by the P6 judge.
\end{promptbox}

\paragraph{Reproducibility note.}
For every LLM call, OR-Space records the model id, temperature, the final
rendered user message, and the raw reply in
\texttt{logs/logs\_\{stage\}/<instance\_id>.txt}. The prompts above are the
\emph{system} side of those records; the \emph{user} side is automatically
constructed from the workspace JSON at runtime, so every prompt in the paper
can be re-assembled from the public artefact plus the released code.

\subsection{Release, Reproducibility, and Responsible Use}
\label{app:release_responsible_use}

\ifarxivversion
\paragraph{Public artefacts.}
The public benchmark package is organized as a benchmark package rather than
as a model checkpoint. It contains the \build, \revise, and \explain workspaces,
split manifests, oracle objectives, solver artefacts, prompt templates,
evaluation rubrics, runner metadata, baseline outputs, dataset and evaluation
cards, a release manifest, and Croissant metadata with Responsible-AI fields.
The arXiv release is intended to be citeable through an immutable public
version tag rather than a moving branch.
\fi

\paragraph{Limitations.}
OR-Space is a synthetic benchmark derived from 100 industrial-style
optimization topologies, so it should not be interpreted as a complete model of
enterprise OR deployments. The workspaces emphasize linear and mixed-integer
programming patterns and may underrepresent nonlinear, stochastic, simulation,
or human-in-the-loop modelling workflows. Solver-backed scoring depends on
solver availability, numerical tolerances, and API familiarity; closed-source
model results may drift as provider endpoints change. \explain scoring combines
exact-match checks with an LLM-judge component, so judge bias and prompt
sensitivity remain possible despite rubric grounding and explicit
hallucination penalties.

\paragraph{Compute and uncertainty.}
\build and \revise submissions run in a CPU-only, network-isolated Docker
environment with a fresh Python interpreter and a 120\,s wall-clock limit per
submission. Gurobi~12.0.1~\cite{gurobi} is used for oracle generation and the
main backend, while solver-robustness tracks rerun selected settings with
COPT~\cite{geCardinalOptimizerCOPT2022}, PuLP/CBC, and HiGHS; the benchmark package records
solver versions, model ids, prompts, decoding settings, token budgets, raw
outputs, and scoring traces. LLM inference for hosted models is performed
through provider APIs, so provider-side hardware is not controlled by the
authors. The main Pass@1 numbers are fixed-split descriptive rates over
$n=100$ instances; one solved instance corresponds to one percentage point, and
a binomial standard error can be read as
$\sqrt{p(1-p)/100}$ for a rate $p$. We therefore interpret small differences
cautiously and emphasize large cross-task gaps and qualitative failure modes.

\paragraph{Licensing and responsible use.}
The seed IndustryOR topologies are cited in the paper and are used under their
published non-commercial research terms; the OR-Space release uses a compatible
research license and does not redistribute proprietary solver binaries or API
credentials. The benchmark contains synthetic workspaces and is not intended to
include personal or sensitive information. Its intended positive use is to
support reproducible evaluation of optimization agents and to expose failures
before deployment. A negative use would be to treat high benchmark performance
as certification for production optimization decisions; the dataset card
therefore states that generated models and solver outputs require independent
inspection before operational use.

\section{Additional Experimental Results}
\label{app:additional_experimental_results}

\subsection{Full Solver-Backend Results}
\label{app:full_solver_backend_results}

This subsection details the complete per-model results across all solver backends used in our robustness analysis. Rows match the model order and task definitions of Table~\ref{tab:main_results}. \build and the three \revise variants report Pass@1 under the objective oracle $\mathcal{M}_{\text{obj}}$, while \explain reports the rubric score $\mathcal{M}_{\text{exp}}$. These results highlight solver sensitivity and support the aggregate comparisons in Table~\ref{tab:solver_comparison}, rather than serving as a separate leaderboard.

\begingroup
\setlength{\LTleft}{\fill}
\setlength{\LTright}{\fill}
\setlength{\tabcolsep}{3pt}
\renewcommand{\arraystretch}{1}

\refstepcounter{table}
\label{tab:appendix_solver_full}

\noindent\makebox[\textwidth][c]{%
\parbox{\textwidth}{\normalsize
Table~\thetable: Full model-wise results under four solver backends on OR-Space
(100 instances per task). \build and the three revise variants report Pass@1
($\mathcal{M}_{\text{obj}}$); \explain reports the rubric mean score in
$[0,100]$. Higher is better for all metric columns; within each solver and
model category, best values are bolded and second-best values are
underlined. Model order and grouping follow Table~\ref{tab:main_results}.}}
\vspace{0.6em}

\scriptsize

\begin{longtable}{llccccc}
\toprule
\textbf{Solver} & \textbf{Model} & \textbf{\build ($\uparrow$)} & \textbf{\revise-all ($\uparrow$)} & \textbf{\revise-model ($\uparrow$)} & \textbf{\revise-code ($\uparrow$)} & \textbf{\explain ($\uparrow$)} \\
\midrule
\endfirsthead
\toprule
\textbf{Solver} & \textbf{Model} & \textbf{\build ($\uparrow$)} & \textbf{\revise-all ($\uparrow$)} & \textbf{\revise-model ($\uparrow$)} & \textbf{\revise-code ($\uparrow$)} & \textbf{\explain ($\uparrow$)} \\
\midrule
\endhead
\bottomrule
\endfoot
\multirow{22}{*}{\texttt{Gurobi}} & \multicolumn{6}{>{\columncolor{SJTUIICGroup}}l}{\emph{Closed-source Models}} \\
& \texttt{gemini-3.1-pro} & \bestval{72/100 (72\%)} & \secondval{91/100 (91\%)} & \secondval{84/100 (84\%)} & \bestval{81/100 (81\%)} & 73.00 \\
& \texttt{gemini-3-flash} & \secondval{68/100 (68\%)} & 52/100 (52\%) & 70/100 (70\%) & 45/100 (45\%) & 53.24 \\
& \texttt{claude-opus-4-6} & 62/100 (62\%) & \bestval{93/100 (93\%)} & 81/100 (81\%) & \secondval{80/100 (80\%)} & 80.20 \\
& \texttt{gpt-5.4} & 59/100 (59\%) & 90/100 (90\%) & \bestval{85/100 (85\%)} & 79/100 (79\%) & \bestval{86.52} \\
& \texttt{gpt-5-mini} & 58/100 (58\%) & 86/100 (86\%) & 80/100 (80\%) & 77/100 (77\%) & \secondval{82.94} \\
& \texttt{claude-sonnet-4.5} & 53/100 (53\%) & 83/100 (83\%) & 70/100 (70\%) & 78/100 (78\%) & 77.58 \\
& \texttt{claude-sonnet-4.5-thinking} & 53/100 (53\%) & 84/100 (84\%) & 73/100 (73\%) & 76/100 (76\%) & 79.63 \\
& \texttt{gpt-5.1} & 53/100 (53\%) & 84/100 (84\%) & 79/100 (79\%) & 71/100 (71\%) & 76.16 \\
& \texttt{gemini-2.5-flash} & 53/100 (53\%) & 67/100 (67\%) & 59/100 (59\%) & 66/100 (66\%) & 13.79 \\
& \texttt{gemini-2.5-pro} & 47/100 (47\%) & 89/100 (89\%) & 71/100 (71\%) & 69/100 (69\%) & 32.22 \\
& \texttt{gpt-4o} & 25/100 (25\%) & 45/100 (45\%) & 22/100 (22\%) & 36/100 (36\%) & 59.36 \\
\cmidrule(lr){2-7}
& \multicolumn{6}{>{\columncolor{SJTUIICGroup}}l}{\emph{Open-source Models}} \\
& \texttt{deepseek-v4-pro} & \bestval{67/100 (67\%)} & \bestval{86/100 (86\%)} & \bestval{82/100 (82\%)} & \bestval{71/100 (71\%)} & 68.87 \\
& \texttt{deepseek-r1-0528} & \secondval{55/100 (55\%)} & 72/100 (72\%) & 70/100 (70\%) & \secondval{65/100 (65\%)} & 44.40 \\
& \texttt{deepseek-v4-flash} & 54/100 (54\%) & \secondval{73/100 (73\%)} & 64/100 (64\%) & 63/100 (63\%) & \secondval{77.77} \\
& \texttt{qwen3-max} & 49/100 (49\%) & 71/100 (71\%) & \secondval{71/100 (71\%)} & 60/100 (60\%) & 74.66 \\
& \texttt{qwen3-32b} & 32/100 (32\%) & 32/100 (32\%) & 24/100 (24\%) & 20/100 (20\%) & 61.32 \\
& \texttt{qwen3-32b-thinking} & 28/100 (28\%) & 31/100 (31\%) & 28/100 (28\%) & 32/100 (32\%) & 61.28 \\
& \texttt{qwen3.5-27b} & 25/100 (25\%) & 40/100 (40\%) & 41/100 (41\%) & 48/100 (48\%) & \bestval{77.88} \\
& \texttt{qwen3-8b} & 22/100 (22\%) & 9/100 (9\%) & 11/100 (11\%) & 12/100 (12\%) & 55.19 \\
& \texttt{qwen3-14b} & 20/100 (20\%) & 18/100 (18\%) & 16/100 (16\%) & 19/100 (19\%) & 61.03 \\
\midrule
\multirow{22}{*}{\texttt{COPT}} & \multicolumn{6}{>{\columncolor{SJTUIICGroup}}l}{\emph{Closed-source Models}} \\
& \texttt{gemini-3.1-pro} & \bestval{73/100 (73\%)} & \secondval{85/100 (85\%)} & 82/100 (82\%) & \secondval{83/100 (83\%)} & 74.89 \\
& \texttt{gemini-3-flash} & 57/100 (57\%) & 80/100 (80\%) & 71/100 (71\%) & 73/100 (73\%) & 58.22 \\
& \texttt{claude-opus-4-6} & \secondval{70/100 (70\%)} & \bestval{89/100 (89\%)} & \bestval{88/100 (88\%)} & \bestval{84/100 (84\%)} & 73.76 \\
& \texttt{gpt-5.4} & 66/100 (66\%) & 83/100 (83\%) & \secondval{86/100 (86\%)} & 80/100 (80\%) & \bestval{81.43} \\
& \texttt{gpt-5-mini} & 8/100 (8\%) & 12/100 (12\%) & 14/100 (14\%) & 16/100 (16\%) & \secondval{77.88} \\
& \texttt{claude-sonnet-4.5} & 52/100 (52\%) & 83/100 (83\%) & 62/100 (62\%) & 78/100 (78\%) & 75.69 \\
& \texttt{claude-sonnet-4.5-thinking} & 57/100 (57\%) & 83/100 (83\%) & 68/100 (68\%) & 78/100 (78\%) & 75.11 \\
& \texttt{gpt-5.1} & 45/100 (45\%) & 55/100 (55\%) & 48/100 (48\%) & 58/100 (58\%) & 74.68 \\
& \texttt{gemini-2.5-flash} & 24/100 (24\%) & 39/100 (39\%) & 18/100 (18\%) & 33/100 (33\%) & 13.06 \\
& \texttt{gemini-2.5-pro} & 45/100 (45\%) & 75/100 (75\%) & 63/100 (63\%) & 77/100 (77\%) & 25.29 \\
& \texttt{gpt-4o} & 21/100 (21\%) & 22/100 (22\%) & 16/100 (16\%) & 31/100 (31\%) & 51.83 \\
\cmidrule(lr){2-7}
& \multicolumn{6}{>{\columncolor{SJTUIICGroup}}l}{\emph{Open-source Models}} \\
& \texttt{deepseek-v4-pro} & 36/100 (36\%) & \bestval{74/100 (74\%)} & 46/100 (46\%) & 49/100 (49\%) & 33.56 \\
& \texttt{deepseek-r1-0528} & \bestval{58/100 (58\%)} & \secondval{68/100 (68\%)} & \secondval{63/100 (63\%)} & \secondval{57/100 (57\%)} & 36.40 \\
& \texttt{deepseek-v4-flash} & 43/100 (43\%) & 48/100 (48\%) & 59/100 (59\%) & \bestval{58/100 (58\%)} & 23.85 \\
& \texttt{qwen3-max} & \secondval{51/100 (51\%)} & 57/100 (57\%) & \bestval{65/100 (65\%)} & \bestval{58/100 (58\%)} & \secondval{68.76} \\
& \texttt{qwen3-32b} & 21/100 (21\%) & 35/100 (35\%) & 25/100 (25\%) & 22/100 (22\%) & 53.28 \\
& \texttt{qwen3-32b-thinking} & 21/100 (21\%) & 31/100 (31\%) & 25/100 (25\%) & 26/100 (26\%) & 56.85 \\
& \texttt{qwen3.5-27b} & 13/100 (13\%) & 24/100 (24\%) & 30/100 (30\%) & 20/100 (20\%) & \bestval{76.14} \\
& \texttt{qwen3-8b} & 1/100 (1\%) & 1/100 (1\%) & 1/100 (1\%) & 0/100 (0\%) & 53.77 \\
& \texttt{qwen3-14b} & 4/100 (4\%) & 3/100 (3\%) & 3/100 (3\%) & 1/100 (1\%) & 58.48 \\
\midrule
\multirow{22}{*}{\texttt{PuLP}} & \multicolumn{6}{>{\columncolor{SJTUIICGroup}}l}{\emph{Closed-source Models}} \\
& \texttt{gemini-3.1-pro} & \bestval{72/100 (72\%)} & \bestval{88/100 (88\%)} & \bestval{81/100 (81\%)} & \bestval{80/100 (80\%)} & 68.06 \\
& \texttt{gemini-3-flash} & 59/100 (59\%) & 80/100 (80\%) & 73/100 (73\%) & 72/100 (72\%) & 60.47 \\
& \texttt{claude-opus-4-6} & 64/100 (64\%) & 84/100 (84\%) & \bestval{81/100 (81\%)} & \secondval{78/100 (78\%)} & 70.59 \\
& \texttt{gpt-5.4} & 60/100 (60\%) & \secondval{85/100 (85\%)} & 75/100 (75\%) & \bestval{80/100 (80\%)} & \bestval{82.33} \\
& \texttt{gpt-5-mini} & \secondval{66/100 (66\%)} & 84/100 (84\%) & \secondval{76/100 (76\%)} & \secondval{79/100 (79\%)} & 73.94 \\
& \texttt{claude-sonnet-4.5} & 51/100 (51\%) & 76/100 (76\%) & 69/100 (69\%) & 73/100 (73\%) & 71.71 \\
& \texttt{claude-sonnet-4.5-thinking} & 55/100 (55\%) & 79/100 (79\%) & 66/100 (66\%) & \secondval{78/100 (78\%)} & \secondval{74.33} \\
& \texttt{gpt-5.1} & 53/100 (53\%) & 80/100 (80\%) & 67/100 (67\%) & 75/100 (75\%) & 73.56 \\
& \texttt{gemini-2.5-flash} & 49/100 (49\%) & 65/100 (65\%) & 53/100 (53\%) & 62/100 (62\%) & 11.37 \\
& \texttt{gemini-2.5-pro} & 48/100 (48\%) & 75/100 (75\%) & 51/100 (51\%) & 62/100 (62\%) & 23.34 \\
& \texttt{gpt-4o} & 23/100 (23\%) & 40/100 (40\%) & 22/100 (22\%) & 43/100 (43\%) & 52.66 \\
\cmidrule(lr){2-7}
& \multicolumn{6}{>{\columncolor{SJTUIICGroup}}l}{\emph{Open-source Models}} \\
& \texttt{deepseek-v4-pro} & \bestval{54/100 (54\%)} & \bestval{75/100 (75\%)} & \bestval{66/100 (66\%)} & \bestval{72/100 (72\%)} & 28.55 \\
& \texttt{deepseek-r1-0528} & \bestval{54/100 (54\%)} & 60/100 (60\%) & 52/100 (52\%) & \secondval{67/100 (67\%)} & 37.24 \\
& \texttt{deepseek-v4-flash} & \secondval{49/100 (49\%)} & \secondval{70/100 (70\%)} & 56/100 (56\%) & 63/100 (63\%) & 22.95 \\
& \texttt{qwen3-max} & \secondval{49/100 (49\%)} & 63/100 (63\%) & \secondval{57/100 (57\%)} & 61/100 (61\%) & \secondval{70.45} \\
& \texttt{qwen3-32b} & 28/100 (28\%) & 32/100 (32\%) & 23/100 (23\%) & 29/100 (29\%) & 57.09 \\
& \texttt{qwen3-32b-thinking} & 30/100 (30\%) & 28/100 (28\%) & 26/100 (26\%) & 33/100 (33\%) & 55.30 \\
& \texttt{qwen3.5-27b} & 41/100 (41\%) & 47/100 (47\%) & \bestval{66/100 (66\%)} & 38/100 (38\%) & \bestval{74.26} \\
& \texttt{qwen3-8b} & 11/100 (11\%) & 8/100 (8\%) & 1/100 (1\%) & 12/100 (12\%) & 53.77 \\
& \texttt{qwen3-14b} & 28/100 (28\%) & 24/100 (24\%) & 10/100 (10\%) & 23/100 (23\%) & 60.21 \\
\midrule
\multirow{22}{*}{\texttt{HiGHS}} & \multicolumn{6}{>{\columncolor{SJTUIICGroup}}l}{\emph{Closed-source Models}} \\
& \texttt{gemini-3.1-pro} & \secondval{66/100 (66\%)} & 76/100 (76\%) & 76/100 (76\%) & 77/100 (77\%) & 57.65 \\
& \texttt{gemini-3-flash} & 49/100 (49\%) & 73/100 (73\%) & 58/100 (58\%) & 70/100 (70\%) & 50.10 \\
& \texttt{claude-opus-4-6} & \bestval{67/100 (67\%)} & \secondval{84/100 (84\%)} & \bestval{87/100 (87\%)} & \bestval{83/100 (83\%)} & 75.25 \\
& \texttt{gpt-5.4} & 59/100 (59\%) & \bestval{85/100 (85\%)} & \secondval{85/100 (85\%)} & \secondval{80/100 (80\%)} & \bestval{82.20} \\
& \texttt{gpt-5-mini} & 64/100 (64\%) & 80/100 (80\%) & 80/100 (80\%) & 72/100 (72\%) & 75.31 \\
& \texttt{claude-sonnet-4.5} & 60/100 (60\%) & 68/100 (68\%) & 66/100 (66\%) & 73/100 (73\%) & 75.57 \\
& \texttt{claude-sonnet-4.5-thinking} & 58/100 (58\%) & 69/100 (69\%) & 65/100 (65\%) & 75/100 (75\%) & 73.86 \\
& \texttt{gpt-5.1} & 49/100 (49\%) & 48/100 (48\%) & 51/100 (51\%) & 49/100 (49\%) & 74.50 \\
& \texttt{gemini-2.5-flash} & 43/100 (43\%) & 32/100 (32\%) & 35/100 (35\%) & 37/100 (37\%) & 12.58 \\
& \texttt{gemini-2.5-pro} & 34/100 (34\%) & 66/100 (66\%) & 53/100 (53\%) & 67/100 (67\%) & 23.29 \\
& \texttt{gpt-4o} & 37/100 (37\%) & 37/100 (37\%) & 34/100 (34\%) & 24/100 (24\%) & 54.42 \\
\cmidrule(lr){2-7}
& \multicolumn{6}{>{\columncolor{SJTUIICGroup}}l}{\emph{Open-source Models}} \\
& \texttt{deepseek-v4-pro} & \bestval{57/100 (57\%)} & 35/100 (35\%) & \bestval{75/100 (75\%)} & \secondval{63/100 (63\%)} & 26.02 \\
& \texttt{deepseek-r1-0528} & 37/100 (37\%) & 34/100 (34\%) & 43/100 (43\%) & 37/100 (37\%) & 37.36 \\
& \texttt{deepseek-v4-flash} & \secondval{54/100 (54\%)} & \bestval{61/100 (61\%)} & \secondval{58/100 (58\%)} & 61/100 (61\%) & 22.15 \\
& \texttt{qwen3-max} & 52/100 (52\%) & \secondval{55/100 (55\%)} & 56/100 (56\%) & \bestval{64/100 (64\%)} & \secondval{68.92} \\
& \texttt{qwen3-32b} & 34/100 (34\%) & 23/100 (23\%) & 25/100 (25\%) & 19/100 (19\%) & 53.68 \\
& \texttt{qwen3-32b-thinking} & 33/100 (33\%) & 22/100 (22\%) & 19/100 (19\%) & 27/100 (27\%) & 52.33 \\
& \texttt{qwen3.5-27b} & 33/100 (33\%) & 31/100 (31\%) & 36/100 (36\%) & 47/100 (47\%) & \bestval{76.41} \\
& \texttt{qwen3-8b} & 7/100 (7\%) & 1/100 (1\%) & 2/100 (2\%) & 2/100 (2\%) & 53.27 \\
& \texttt{qwen3-14b} & 14/100 (14\%) & 15/100 (15\%) & 9/100 (9\%) & 16/100 (16\%) & 59.13 \\
\end{longtable}
\endgroup

The full table is intentionally exhaustive: it preserves model-level variation
within each solver track, including cases where a solver interface changes a
model's failure mode rather than only its objective value. The main text reports
the headline filesystem, solver, and \revise-context summaries derived from
these runs. The following subsections therefore avoid repeating the main-text
tables and instead report compact delta views and additional model-level
diagnostics.

\subsection{Filesystem Interface Delta Details}
\label{appendix:fs_vs_flat}

The main text reports the full filesystem-vs-flat table. To make the interface
effect easier to inspect in the appendix, Table~\ref{tab:appendix_fs_flat_delta}
reports only the per-model deltas, defined as filesystem score minus flat-prompt
score under the same task and model. Negative values therefore indicate a
filesystem penalty.

\begin{table}[H]
\centering
\caption{Per-model filesystem effect on the controlled six-model subset.
Entries are filesystem minus flat-prompt scores. \build and
\revisecode are Pass@1 percentage-point differences; \explain is a
Rubric-score difference. Negative values indicate that the filesystem interface
reduced performance relative to the flat prompt.}
\label{tab:appendix_fs_flat_delta}
\footnotesize
\setlength{\tabcolsep}{6pt}
\renewcommand{\arraystretch}{1.04}
\begin{tabular}{lccc}
\toprule
\textbf{Model} & \textbf{\build $\Delta$} & \textbf{\revisecode $\Delta$} & \textbf{\explain $\Delta$} \\
\midrule
\texttt{gemini-3.1-pro}    & $0$   & $-13$ & $+0.45$ \\
\texttt{deepseek-v4-pro}   & $-3$  & $-14$ & $+0.11$ \\
\texttt{gpt-5.4}           & $-7$  & $-10$ & $+0.35$ \\
\texttt{claude-sonnet-4.5} & $-9$  & $-7$  & $+1.90$ \\
\texttt{qwen3-max}         & $-11$ & $-15$ & $-2.53$ \\
\texttt{gpt-4o}            & $-19$ & $-14$ & $+1.81$ \\
\midrule
\textit{Average}           & $-8.2$ & $-12.2$ & $+0.3$ \\
\bottomrule
\end{tabular}
\end{table}

The delta view shows that the filesystem penalty is broad but not uniform. It
is largest for \texttt{gpt-4o} on \build ($-19$\,pp) and for
\texttt{qwen3-max} on \revisecode ($-15$\,pp), while
\texttt{gemini-3.1-pro} is insensitive to the interface for \build but loses
$13$\,pp on \revisecode. This supports the main-text interpretation
that file discovery, path-correct code generation, and schema inspection are
part of the task rather than superficial formatting. \explain remains much
closer across interfaces because flattening removes navigation but not the
need to reconcile documents, data, code, and solver records.

\subsection{Solver-Backend Sensitivity Details}
\label{appendix:solver_comparison}

The main text reports solver-level macro averages. The full table above shows
why that aggregate view should be read with caution: solver effects are highly
model-specific, and a model can be stable under one backend while failing under
another. Table~\ref{tab:appendix_solver_sensitivity_examples} lists
representative \build rows that illustrate this spread.

\begin{table}[!htbp]
\centering
\caption{Representative model-level \build sensitivity across solver backends.
Entries are \build Pass@1 (\%). Range is the max--min spread across the four
solver tracks for the same model.}
\label{tab:appendix_solver_sensitivity_examples}
\footnotesize
\setlength{\tabcolsep}{5pt}
\renewcommand{\arraystretch}{1.04}
\begin{tabular}{lccccc}
\toprule
\textbf{Model} & \textbf{Gurobi} & \textbf{COPT} & \textbf{PuLP} & \textbf{HiGHS} & \textbf{Range} \\
\midrule
\texttt{gpt-5-mini}        & 58 & 8  & 66 & 64 & 58 \\
\texttt{deepseek-v4-pro}   & 67 & 36 & 54 & 57 & 31 \\
\texttt{qwen3-8b}          & 22 & 1  & 11 & 7  & 21 \\
\texttt{gemini-3.1-pro}    & 72 & 73 & 72 & 66 & 7  \\
\texttt{claude-opus-4-6}   & 62 & 70 & 64 & 67 & 8  \\
\bottomrule
\end{tabular}
\end{table}

Two implications follow. First, backend robustness is not monotone in model
scale: \texttt{gpt-5-mini} is competitive under Gurobi, PuLP, and HiGHS, but
drops sharply under COPT in this track. Secondly, stable models are not always
the highest-scoring models; \texttt{gemini-3.1-pro} and
\texttt{claude-opus-4-6} have small solver ranges, while
\texttt{deepseek-v4-pro} has a much larger spread. This is why the main text
reports both the default Gurobi track and the four-backend aggregate rather
than treating any single backend as a complete evaluation.

\subsection{\revise Context Model-Level Details}
\label{appendix:revise_context_ablation}

The main text reports the solver-level \revise context ablation for all completed
model rows and for the top-5 subset. Table~\ref{tab:appendix_revise_gurobi_delta}
adds a model-level Gurobi delta view, which makes clear that the same context
choice can help one model and hurt another.

\begin{table}[!htbp]
\centering
\caption{Model-level Gurobi \revise context deltas. Code, All, and Model are
Pass@1 (\%) for \revisecode, \reviseall, and
\revisemodel; deltas are relative to \revisecode.}
\label{tab:appendix_revise_gurobi_delta}
\footnotesize
\setlength{\tabcolsep}{4pt}
\renewcommand{\arraystretch}{1.04}
\begin{tabular}{lccccc}
\toprule
\textbf{Model} & \textbf{Code} & \textbf{All} & \textbf{Model} & \textbf{All $-$ Code} & \textbf{Model $-$ Code} \\
\midrule
\texttt{gemini-3.1-pro}          & 81 & 91 & 84 & $+10$ & $+3$ \\
\texttt{gemini-3-flash}          & 45 & 52 & 70 & $+7$  & $+25$ \\
\texttt{claude-opus-4-6}         & 80 & 93 & 81 & $+13$ & $+1$ \\
\texttt{gpt-5.4}                 & 79 & 90 & 85 & $+11$ & $+6$ \\
\texttt{gemini-2.5-pro}          & 69 & 89 & 71 & $+20$ & $+2$ \\
\texttt{deepseek-v4-pro}         & 71 & 86 & 82 & $+15$ & $+11$ \\
\texttt{qwen3-max}               & 60 & 71 & 71 & $+11$ & $+11$ \\
\texttt{qwen3.5-27b}             & 48 & 40 & 41 & $-8$  & $-7$ \\
\texttt{qwen3-32b-thinking}      & 32 & 31 & 28 & $-1$  & $-4$ \\
\bottomrule
\end{tabular}
\end{table}

The model-level deltas refine the aggregate conclusion. Adding the formulation
on top of code is broadly helpful for strong Gurobi-track models, but the size
of the all-context gain varies substantially, from $+7$\,pp for
\texttt{gemini-3-flash} to $+20$\,pp for \texttt{gemini-2.5-pro} among the
rows shown. The formulation-only setting can also rescue models that are
harmed by heuristic code, as in \texttt{gemini-3-flash} ($+25$\,pp relative to
code), while weakening others such as \texttt{qwen3.5-27b}. This supports the
claim that code and formulation are complementary signals rather than
interchangeable inputs.

\subsection{Failure Details and Case Studies}
\label{app:failure_details}

Across all $20 \times 100$ \revise trials, $19.1\%$ of submissions are
\wrongvalue (executable script, wrong objective), $18.2\%$ are
\runtimeerror, $3.1\%$ are \typeerror, and $2.8\%$ are
\texttt{module\_not\_found}. The \texttt{module\_not\_found} rate rises $7\times$
from \build ($0.4\%$) to \revise ($2.8\%$), often because agents inherit
non-solver imports such as \texttt{scipy} or \texttt{numpy} from the heuristic.
\nameerror similarly rises from $0.2\%$ to $2.2\%$ when agents reuse
heuristic variable names without redeclaring them as solver variables.

Table~\ref{tab:appendix_failure_cases} expands the main-text case study with
additional representative \build failures from the same manual audit. The first
row corresponds to \texttt{IndustryOR\_2}, discussed in
Section~\ref{subsec:failure_case_study}; the remaining rows illustrate
additional workspace-specific failure types.

\begin{table}[H]
\centering
\caption{Representative \build failures for \texttt{gemini-3.1-pro}. These
cases extend the main-text case study by showing several distinct
workspace-specific error types.}
\label{tab:appendix_failure_cases}
\scriptsize
\setlength{\tabcolsep}{3pt}
\renewcommand{\arraystretch}{1.08}
\begin{tabular}{p{0.13\textwidth}p{0.18\textwidth}p{0.23\textwidth}p{0.36\textwidth}}
\toprule
\textbf{Instance} & \textbf{Observed result} & \textbf{Workspace evidence} & \textbf{Root cause} \\
\midrule
\texttt{IndustryOR\_2} & Wrong objective value: 25 vs.\ reference 125 &
\texttt{a1}=10, \texttt{a2}=15, and \texttt{training\_capacity\_per\_jet}=5
define trained-pilot capacity. &
The agent introduced combat-jet variables and optimized \texttt{C1 + C2},
mapping training capacity to the wrong physical quantity. \\
\addlinespace[0.2em]
\texttt{IndustryOR\_9} & Runtime parser error before model construction &
\texttt{general\_parameters.csv} includes free-text business descriptions with
commas in the context column. &
The agent loaded the full CSV with \texttt{pd.read\_csv}; the descriptive
column triggered a Pandas C-parser error instead of being ignored or parsed
robustly. \\
\addlinespace[0.2em]
\texttt{IndustryOR\_27} & Runtime \textsc{KeyError}: \texttt{unit\_price\_product\_i} &
The CSV keys use Roman numerals with case-sensitive suffixes, e.g.,
\texttt{unit\_price\_product\_I}. &
The agent normalized product names with \texttt{p.lower()}, creating keys that
do not exist in the parameter file. \\
\addlinespace[0.2em]
\texttt{IndustryOR\_18} & Wrong objective value: 255000 vs.\ reference 5 &
The document asks the factory to fully utilize time, minimize overtime, and
meet fabric sales targets. &
The agent converted the task into profit maximization over fabric output,
changing the business objective rather than only encoding the stated
requirements. \\
\bottomrule
\end{tabular}
\end{table}

\paragraph{Wrong business quantity.}
In \texttt{IndustryOR\_2}, the workspace describes a two-year fighter-pilot
training plan. The data fields specify \texttt{a1}=10 fighter jets in year 1,
\texttt{a2}=15 fighter jets in year 2, and
\texttt{training\_capacity\_per\_jet}=5 pilots per year; the reference logic
therefore counts $5(10+15)=125$ trained pilots. The generated script instead
introduced \texttt{T1}, \texttt{C1}, \texttt{T2}, and \texttt{C2}, constrained
\texttt{C2 <= training\_capacity\_per\_jet * T1}, and maximized
\texttt{C1 + C2}. The model solved to optimality, but it optimized combat jets
rather than the trained-pilot quantity encoded by the workspace.

\paragraph{CSV parsing and schema grounding.}
\texttt{IndustryOR\_9} and \texttt{IndustryOR\_27} illustrate two distinct data
access failures. In \texttt{IndustryOR\_9}, the relevant numeric parameters are
in the first two columns of \texttt{general\_parameters}\allowbreak\texttt{.csv}, while the context-description column contains natural-language rules such as ``If Haus Toys manufactures boats, they will also manufacture airplanes.'' Because that
text contains commas, the generated \texttt{pd.read\_csv} call failed with
\texttt{Expected 4 fields in line 4, saw 5}. In \texttt{IndustryOR\_27}, the
script successfully read the CSV but constructed nonexistent lowercase keys
such as \texttt{unit\_price\_product\_i}; the actual parameter keys preserve
Roman-numeral capitalization, e.g., \texttt{unit\_price\_product\_I}. These
cases are not mathematical modelling errors in the narrow sense. They are
workspace-grounding failures caused by brittle assumptions about file format
and exact schema names.

\paragraph{Objective semantics.}
\texttt{IndustryOR\_18} shows a different failure type. The business document
states that the textile factory should meet minimum sales targets, fully use
weekly production time, and minimize overtime. The generated model imposed the
minimum sales constraints but then maximized
\texttt{curtain\_fabric\_profit * x\_curtain + clothing\_fabric\_profit *
x\_clothing}. It returned an \optimalstatus objective value of 255000,
whereas the reference objective is 5. The issue is not a missing constraint or
solver crash; it is an objective-alignment error in which the agent used a
plausible business metric available in the CSV while ignoring the operational
goal stated in the document.

\section{Supplementary Dataset Statistics and Visualizations}
\label{app:supp_plots}
In this subsection, we provide additional descriptive statistics of the OR-Space dataset and a granular breakdown of agent performance across different task phases.

\begin{figure}[H]
  \centering
  \includegraphics[width=1\textwidth]{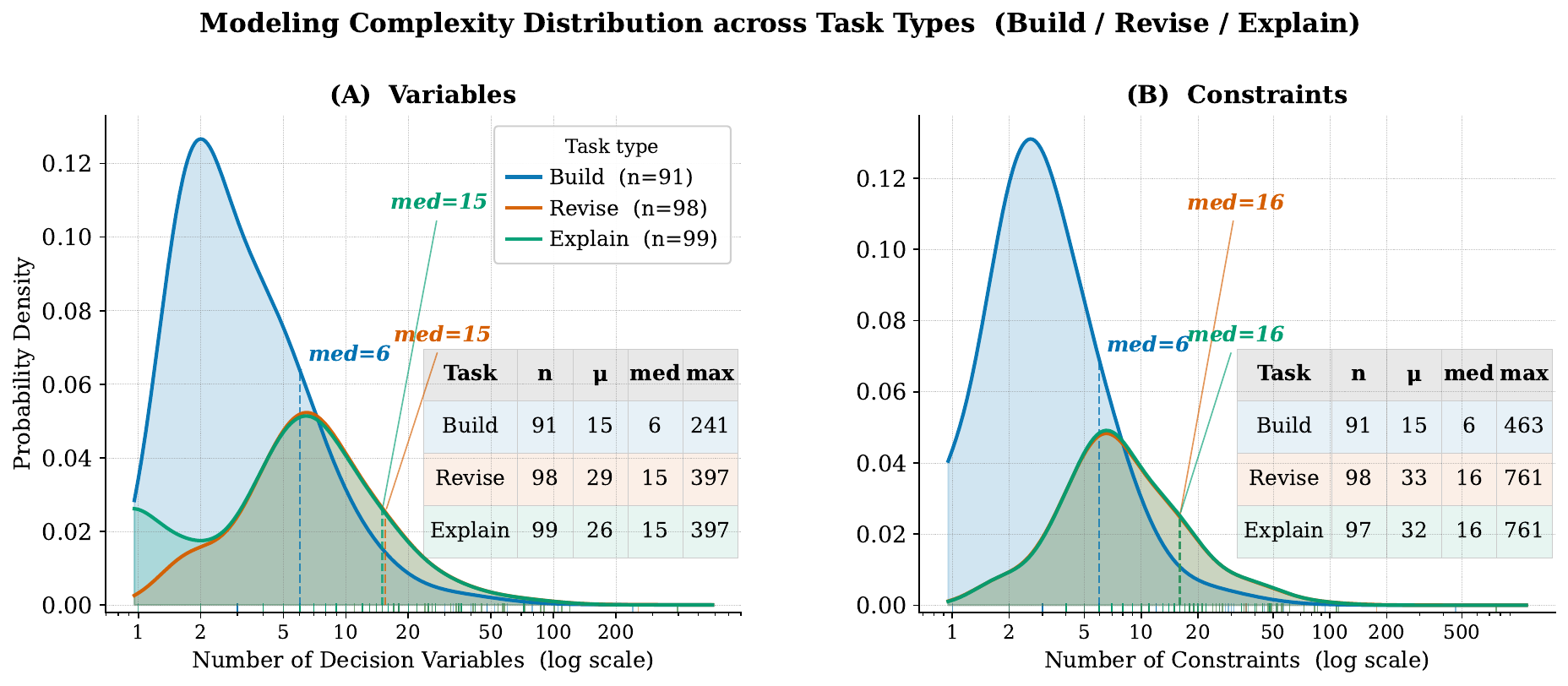} 
  \caption{Kernel Density Estimation (KDE) of constraint and decision variable distributions across the OR-Space dataset, illustrating both the scale of optimization logic and the dimensionality of the problem instances.}
  \label{fig:app_kde_complexity}
\end{figure}

\begin{figure}[H]
  \centering
  \includegraphics[width=1\textwidth]{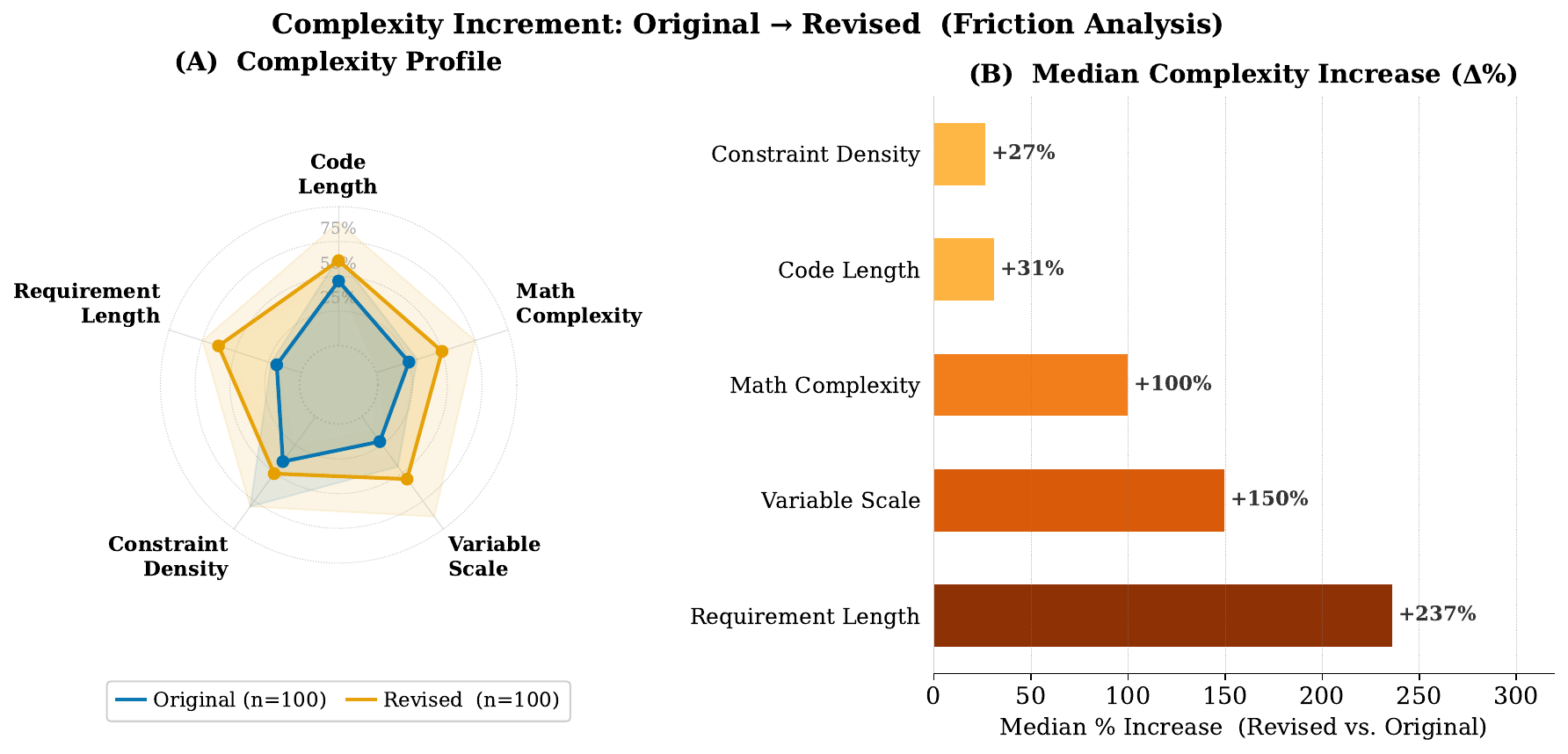}
  \caption{Multi-dimensional complexity analysis of the OR-Space dataset. The radar plots compare various metrics across domains and highlight the logical increments introduced between the \build and \revise tasks.}
  \label{fig:app_radar_complexity}
\end{figure}

\begin{figure}[H]
  \centering
  \includegraphics[width=0.75\textwidth]{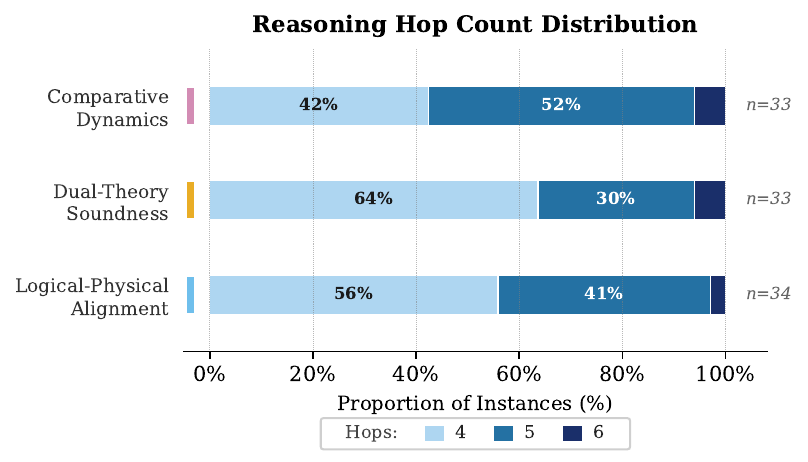}
  \caption{Detailed breakdown of reasoning hops and multi-step logic required for successful task completion in the \explain phase.}
  \label{fig:app_hops}
\end{figure}

\section{\explain Evaluation Rubric}
\label{appendix:rubric}

To ensure a rigorous and objective assessment of the open-ended, short-answer responses in the \textbf{\explain} task, we implement an \textit{LLM-as-judge} protocol. Given that these tasks require multi-hop reasoning across heterogeneous artifacts—including business documents ($\docset$), parameters ($\paramset$), and solver records or diagnostic states—a simple keyword-matching approach is insufficient.

We provide the judge (e.g., \texttt{GPT-4o}) with the ground-truth solver
output and a structured five-dimensional rubric. Each response receives a
score in $[0,100]$ before clipping, with a hallucination penalty applied for
unsupported constraints, parameters, or solver claims.

\begin{table}[H]
  \caption{Scoring rubric for the \explain task. Positive dimensions sum to 100 points; hallucinated or unsupported claims can subtract up to 12 points before the final score is clipped to the reported range.}
  \label{tab:explain-rubric}
  \centering
  \small
  \setlength{\tabcolsep}{4pt}
  \renewcommand{\arraystretch}{1.0}
  \begin{tabular}{@{}lp{8.4cm}r@{}}
    \toprule
    \textbf{Dimension} & \textbf{Evaluation Criteria} & \textbf{Points} \\
    \midrule
    Exact Coverage & Includes the required atomic facts, including variable names, constraint names, CSV columns, units, and numeric values. & 35 \\
    \addlinespace[0.15em]
    Reasoning & Correctly connects the facts through the relevant optimization logic, such as binding constraints, slack, objective changes, or revision effects. & 35 \\
    \addlinespace[0.15em]
    Grounding & Grounds the explanation in the provided workspace artifacts rather than generic OR knowledge or unstated assumptions. & 20 \\
    \addlinespace[0.15em]
    Answer Quality & Gives a concise, coherent answer that directly addresses the question and uses clear terminology. & 10 \\
    \addlinespace[0.15em]
    Hallucination Penalty & Subtracts credit for unsupported constraints, parameters, solver facts, or causal claims not present in the workspace. & up to $-12$ \\
    \bottomrule
  \end{tabular}
\end{table}

\subsubsection{Scoring Guidelines}
The judge is instructed to penalize hallucinated constraints or parameters not present in the workspace. The final score is based on the weighted rubric above, where:
\begin{itemize}
    \setlength{\topsep}{0.2em}
    \setlength{\itemsep}{0pt}
    \setlength{\parsep}{0pt}
    \setlength{\parskip}{0pt}
    \item \textbf{90--100 (Excellent):} The explanation covers the required facts, gives the correct causal or mathematical reasoning, and grounds the answer in the relevant workspace artifacts without hallucination.
    \item \textbf{60--89 (Good):} The core answer is correct, but it may omit a supporting fact, give an incomplete reasoning chain, or provide weaker artifact grounding.
    \item \textbf{30--59 (Fair):} The response identifies part of the answer but gives vague, incomplete, or partially incorrect mathematical justification.
    \item \textbf{0--29 (Poor):} The response is irrelevant, contradicts the solver output, or fails to perform basic cross-file mapping.
\end{itemize}

\end{document}